# Deep encoding of etymological information in TEI

Jack Bowers, OEAW & Inria
Laurent Romary, Inria & BBAW & CMB

## Abstract

In this paper we provide a systematic and comprehensive set of modeling principles for representing etymological data in digital dictionaries using TEI. The purpose is to integrate in one coherent framework both digital representations of legacy dictionaries and born-digital lexical databases that are constructed manually or semi-automatically.

We provide examples from many different types of etymological phenomena from traditional lexicographic practice, as well as analytical approaches from functional and cognitive linguistics such as metaphor, metonymy and grammaticalization, which in many lexicographical and formal linguistic circles have not often been treated as truly etymological in nature, and have thus been largely left out of etymological dictionaries.

In order to fully and accurately express the phenomena and their structures, we have made several proposals for expanding and amending some aspects of the existing TEI framework.

Finally, with reference to both synchronic and diachronic data, we also demonstrate how encoders may integrate semantic web/linked open data information resources into TEI dictionaries as a basis for the sense, and/or the semantic domain of an entry and/or an etymon.

## 1. Introduction

This paper aims to provide a comprehensive modeling and representation of etymological data in digital dictionaries. The purpose is to integrate in one coherent framework both digital representations of legacy dictionaries and born-digital lexical databases that are constructed manually or semi-automatically. We propose a systematic and coherent set of modeling principles for a variety of etymological phenomena that may contribute to the creation of a continuum between existing and future lexical constructs, so that anyone interested in tracing the history of words and their meanings will be able to seamlessly query lexical resources.

Instead of designing an ad hoc model and representation language for digital etymological data, we will focus on identifying all the possibilities offered by the TEI Guidelines for the representation of lexical information. This will lead us to systematize some existing constructs offered by the existing TEI framework, in particular the use of citation (<cit>) for representing etymons in replacement to <mentioned> and referencing constructs(<pRef> and <oRef>) for linking etymological information to existing or putative lexical entries. We also suggest some amendments to the TEI guidelines that may improve the representation of etymological information, as well as lexical entries at large (for instance, deprecation of <oVar> and <pVar>)[1].

---

[1] Some of these amendments have been validated by the TEI council at the time of publication of this paper.

Since its initial design in the 1990's (Ide and Véronis 1994; Ide and Véronis 1995), the TEI "Dictionaries"[2] chapter has been the basis for a large number of dictionary projects. It has shown its capacity to take into account a variety of perspectives on lexical content, whether one wants to closely follow the original structure of the source material (so called *editorial view*), or abstract away from it to go closer to a real lexical database (*lexical view*). This has led to quite an important body of literature (Erjavec, Tufis, and Varadi 1999;Budin, Majewski, and Moerth 2012; Rennie 2000; Bański and Wójtowicz 2009; and Fomin and Toner 2006, to cite a few);most of these papers have been focused on presenting the general architecture of lexical entries in the corresponding dictionary projects, and on describing the way the various TEI elements have been set up and usedover the course of the editorial workflow.

Concerning etymological description on the theoretical level within the field of linguistics, as we shall see, phenomenasuch as metaphor, metonymy, or grammaticalization are very well established, particularly within cognitive linguistics.Howeverwe know of no attempts to represent such processes within any lexical markup systems. Additionally, with regards to the theoretical background, very little has been written on the corresponding digital models when such information is being integrated in a lexical database. This is why we will mainly position our work as an elaboration upon the seminal proposals of Salmon-Alt (2006), which represent a unique set of approaches to data modeling for etymological information.

Finally, though not the primary focus of our paper, we present herein examples of how encoders may make use of linked open data URI's[3] in defining the semantics (sense and/or domain) of a lexical entry; this issue has been discussed recently by Schopper, Bowers and Wandl-Vogt (2015). The integration of the burgeoning resources of the semantic web with TEI represents a step towards a model of digital lexicography which enables conceptual semantics to play a more prominent role in the representation of linguistic content by grounding such information in the ever growing networks of ontological knowledge bases.

## 2. A quick overview of the TEI recommendations for dictionaries

The representation of lexical information is obviously just one of many types of textual forms that are covered by the wide scope of the TEI Guidelines. As such, a dictionary represented in TEI follows all the basic assumptions concerning the general structure of TEI conformant documents. In particular, all metadata elements related to the identification of the sources used in the document, the various responsibilities in its digital encoding, as well as the possible conditions of publication and re-use can be all described within the TEI header (<teiHeader> element), which is a mandatory component of all TEI documents. In the same way, the actual lexical content of a dictionary document expressed in TEI can be further structured at any depth using the generic division (<div>) mechanisms. As a whole, the structural divisions of TEI dictionary entries and their component elements can be seen as analogous to any other type of structured subsection (title, section headers, paragraph, etc.) that may occur in a document containing prose[4].

Besides generic textual constructs, the TEI Guidelines provide a variety of elements to represent dictionary entries, including a general-purpose <entry> element for structured content, a specific <entryFree> element to provide a flat representation, for instance in the course of a digitization workflow, and a <superEntry> container to group together homonyms. Over the course of this paper we will focus on the <entry> element, whose organisation reflects a standard semasiological model of lexical content[5]. Indeed, the <entry> element is mainly organized around two sub-components:

- a <form> element contains the description of the phonetic, orthographic and morphological characteristics of the head-word as well as its possible inflections. This element may for instance contain further grammatical constraints (<gramGrp> element);
- one or more <sense> elements that group together all descriptions related to the various senses that can be associated with the headword. A variety of further descriptors are available in <sense> to provide such information as a definition (<def>), examples or translations (<cit>), various grammatical (<gramGrp>) or usage (<usg>) constraints, and of course etymological information (<etym>).

---

[2] Originally named "Print dictionaries" before it made an appropriate digital turn to cover lexical resources at large.
[3] Uniform Resource Identifier: Naming mechanism to identify a resource on the Internet in a univocal way.
[4] For non expert readers interested in having a quick overview of general encoding possibilities offered by the TEI Guidelines, we recommend looking at the TEI by example initiative: http://www.teibyexample.org, or Romary, 2009.

The TEI "Dictionaries" chapter also provides various mechanisms for cross-referencing entries to other components of a dictionary. In particular, we will see in this paper how we can make use of references to the orthographic form (<oRef>) or pronunciation (<pRef>) of a headword within an etymological description.

When using the <entry> element, it is particularly important to identify the language information attached to any descriptive element within such representations; in particular, an encoder needs to be able to clearly state the *object language* of the entry as a whole (the language about which the entry provides a lexical description) and the various *working languages* (the languages in which various descriptive objects such as definitions, notes, etymons, etc. are expressed). To this purpose, in compliance with, for instance, ISO standard 16642[6] for terminological data, we recommend using the @xml:lang attribute as follows: @xml:lang is a mandatory attribute for each <entry> and indicates the *object language* of the whole entry. When not superseded by other indication further down in the entry structure it also states the *working language* for all descendant elements within it, when appropriate (i.e. for textual content). When the working language differs locally from the object language of the entry (e.g. a definition expressed in another language than the one being described), a new @xml:lang attribute may be attached to the corresponding element.

The representation of etymological information, whether at entry level or for a specific sense, relies on the <etym> element, which we will elaborate upon as we tackle specific phenomena. So far, <etym> has been used as a flat construct where relevant information concerning language (<lang>), etymon (<mentioned>), or source (<bibl>) for instance would be simply marked up in the flow of a textual etymological description. The purpose of this paper is to deepen the possible usage of <etym> and systematize the way specific phenomena can be represented.

# 3. Past Treatment of Etymological Markup

None of the previous attempts to either create a lasting, well formatted digital corpus of etymological data, or to establish a widely adopted set of recommendations for encoding such information have ultimately been very successful. Many of the projects which attempted to create such resources have seen the same fate as so many others in the humanities (despite their stated goals of following best practices for interoperability); such problems include: obsolescence of formatting and/or encoding scheme, abandonment of project, websites no longer existing, broken links, incompatible software, etc.

---

[5] See Romary and Witt, 2014 for an overview of onomasiological and semasiological models, Lemnitzer et al. 2013 and Romary and Wegstein 2012 for an in-depth analysis of the TEI dictionary model, and Romary 2013 for a discussion of the relation between the TEI dictionary model and the ISO 24613 (LMF) standard.
[6] ISO 16642:2003 Computer applications in terminology -- Terminological markup framework, see also Romary, 2001.

However, it is useful to review a few publications (and data where possible) that have led to our current understanding of a generic way to represent (digital) etymological information, as it helps us to establish an understanding of key questions, challenges and issues that the authors encountered, as well as to recognize what people may be looking for in undertaking such projects. To this end, we will review four main references that have either paved the way for the current status quo in the TEI Guidelines or directly influenced our own understanding of the Guidelines and how they should evolve.

A major milestone in the digital dictionary era is probably the work by Amsler and Tompa (1988), which, together with the unifying contributions of Ide and Véronis (1994), led to the earlier TEI "Print dictionaries" chapter. Focusing here on their contribution to Etymology, we can see (cf. Example 1) how they have introduced a highly structured model based on etymons and links implemented as an SGML[7] DTD[8]. The underlying model is clearly based upon a graph of etymons (<etymon>) connected with relations (<rel>), forming a more global etymological tree (reflected by the <es>, etymological segment, element).

<span style="color:#4472C4">**Example 1: etymological representation from Tompa (1988)**</span>

```
<E>
<es>
<etymon lang=ME>appel</etymon></es>
<es>
<rel>fr.</rel>
<etymon lang=OE>æppel</etymon></es>
<es>
<rel>akin to</rel><eu>
<etymon lang=OHG>apful</etymon>
<deftext>apple</deftext></eu><eu>
<etymon lang=OSlav>abl˘ko</etymon></eu></es></E>
```

These next two examples were early attempts to build corpora capable of representing etymological data using standards.

An early pre-XML application of the TEI Guidelines to systematically record etymological information can be found in Good and Sprouse (2000). The work has been carried out in the context of the Comparative Bantu Online Dictionary (CBOLD), a complex database for multiple Bantu languages, and used the SGML P3 edition of the TEI Guidelines[9]. The content corresponds to the digitization of existing print dictionaries and word lists, and the authors marked up these texts according to the *TEI Guidelines*, with someadded tags to the standard set. Since the project had to deal extensively with etymological information, it used <etym> with refined recommendation to link etymons (marked up as <xref>) to a list of reconstructed historical forms.

In the same vein, Jacobson and Michailovsky (2002) used an even simpler approach for their etymological references within a TEI-based encoding of their lexical data. Instead of implementing <etym>, they make a plain use of the generic <ptr> element, typed as "cfetym", to point to other entries in their dictionary that may be seen as etymological sources.

---

[7] Standard Generalized Markup Language, ISO standard ISO 8879 published in 1986, which is the direct ancestor of XML.
[8] Document Type Definition, the grammar of an SGML document.
[9] The P4 edition of the TEI Guidelines, which was completely based upon XML, was published in 2001.

## 3.1 Crist (2005)[10]

In this paper, Crist provides someanalyses of approaches and a correspondingly precise set of principles to be applied to the Germanic Lexicon Project which was a collection of dictionaries of various Germanic languages whose copyright had expired. The ultimate goal for markup formatting was TEI, but (for reasons unknown) this was apparently never achieved. Notably, Crist mentions the likely need to extend the guidelines in the area of etymology due to the fact that the <etym> element lacks the means of precisely encoding etymological relationships between entries and forms.Key components of this work were the following:

- XML markup of some of the data, while other portions remain as plain text, or simply image scans of the originals
- formal interrelations among all of the words in an etymology; those specified are: *cognation, inheritance, borrowing*;
- use of attribute inheritance for the nodes in the data structure as per Ide et. al (2000);
- proposal for the system to require no privileged frame of reference, which would allow data to follow one of three formats;

The following excerpt is from the paper, assuming that the structures represent the place in the XML hierarchy in which each data type would occur.

Example 2 (Crist 2005): abstracted model of etymological description

1. (From the vantage point of Modern English) Modern English *stone* is a reflex of Old English *stān,* which is a reflex of Proto-Germanic *stainaz:*

```
word
form: stone
language: Modern English
etymon
word
form:       stān
language: Old English
etymon
word
form: stainaz
language: Proto-Germanic
attested: no
```

2. (From the vantage point of Old English) Old English *stān* is an etymon of Modern English *stone*, and is also a reflex of Proto- Germanic *stainaz:*

```
        word
                form:       stān
                language: Old English
                        reflex
                        word
                                        form: stone
                                        language: Modern English
                        etymon
                        word
                                        form: stainaz
                                        language: Proto-Germanic
                                        attested: no
```

3. (From the vantage point of Proto-Germanic) Proto-Germanic *stainaz* is an etymon of Old English *stān,* which is an etymon of Modern English *stone:*

```
word
      form: stainaz
      language: Proto-Germanic
      attested: no
      reflex
            word
                  form:       stān
                  language: Old English
                        form: stone
                        language: Modern English
```

Crist (2005) outlines a typology of the treatment of etymological markup at the time, while the adoption of standards and the field of digital humanities and lexicography in particular have been steadily gaining momentum, with regards to etymological markup, this typology remains fairly valid. The classifications are as follows:

(Type I) *Markup schemes which make no provision for etymological data;* (the majority of lexical markup systems);

(Type II) *Markup schemes where etymological data is delimited as such, but is treated as unstructured prose;* (included in this is TEI; points out need for further structure, possible re-use of structures from other sections of TEI specifications or the dictionary chapter specifically);

(Type III)Markup schemes where the mathematical relationships recognized in historical/comparative linguistics are somehow embodied in the markup system in (semi-)machine-readable form. These systems according toCrist (2005), '*make some provision for the formal encoding of etymological relationships between words*'.

### 3.2 Salmon-Alt (2006)

The most significant attempt at devising this kind of dynamic system of etymological markup was that of Salmon-Alt (2006). While etymology is not addressed in the LMF (ISO 24613:2008) standard[11], Salmon-Alt (2006) made an attempt to develop an extension of the model for the encoding of etymological markup. The extension module allows for the integration and linking of the etymological information of an entry with the synchronic data and any classifications of a given word/entry within the core module of LMF.

> *Ourmodel is based on the overall hypothesis that etymological data might be thought of as a lexical network, i.e. a graph, whose nodes are lexical units (located in space and time) and whose arcs are typed etymological relations. (Salmon-Alt 2006, 3)*

The scope of the model was limited to semasiological organizational principles for single lexical entries, such as those outlined in the TEI P5 Guidelines and did not attempt to support the approaches of many traditional etymological dictionaries in which the structural principles and contents vary significantly from one another.

In laying the conceptual and functional basis for the approach, Salmon-Alt defines etymology proper as concerning the origin and evolution of a lexeme before its entry into the lexiconof a given language, as it is materialized by one or more etymons.

The extension's LMF data structure the diagram of the metamodel from the paper is shown below.

---

[11] In the context of the ongoing revision of the LMF document as a multipart standard, there is now provision for a specific part on diachrony and etymology.

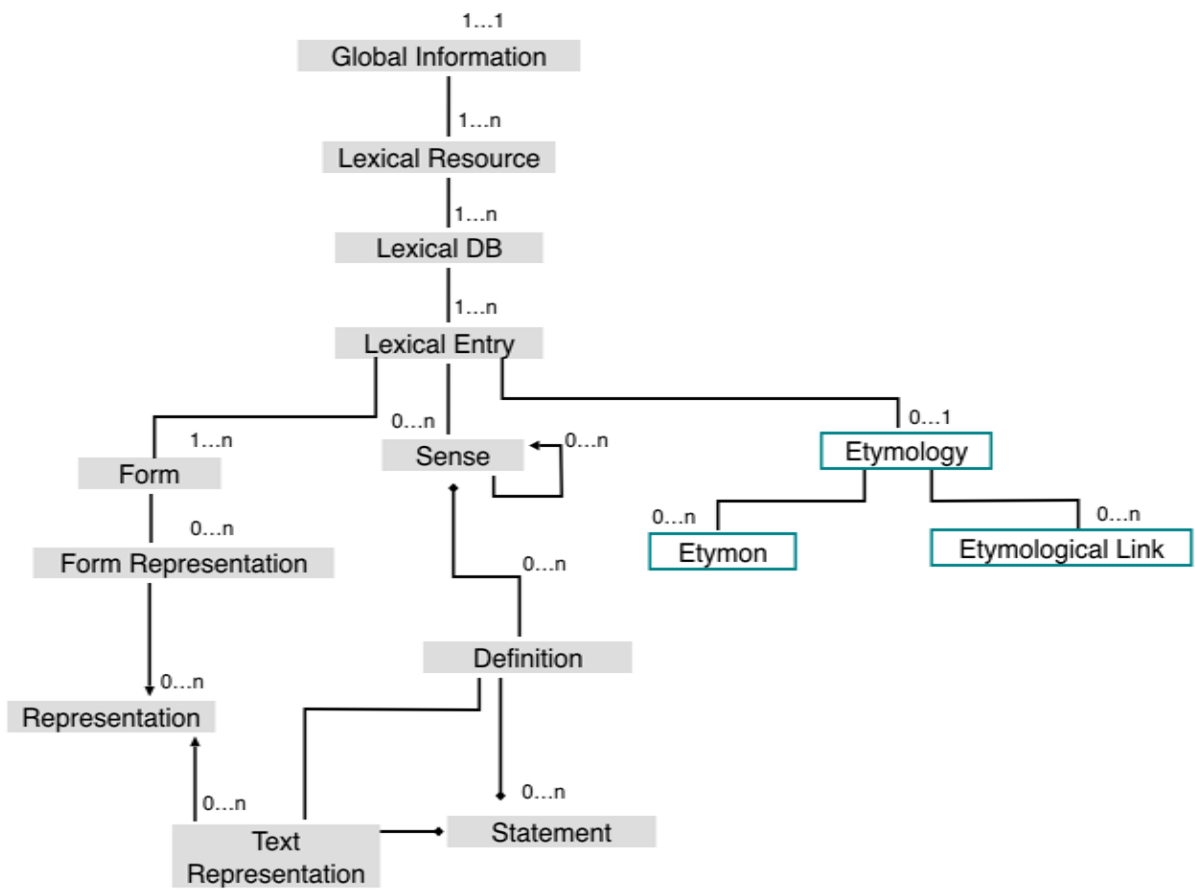

This diagram is reflected in the XML representation suggested by Salmon-Alt (2006)[12], with two dedicated elements: *<etymon> and <etymologicalLink>* which each contain a specific set of information. They are defined as follows:

*Etymon: <etymon>*

The basis for describing and encoding etymons in the model is parallel to that of synchronic lexical entries, specifically, they are characterized by: language (@*xml:lang*), the linguistic form(s) (<form>), orthographic (<orth>), and/or phonetic, sense (<sense>), gloss (<glose>), grammatical classification (<pos>),and inflectional information (*if applicable*).

Additionally within the *etymon* portion of the markup are the optional *etymological notes*, which serve as a kind of '*et cetera*' section where one can include other relevant information about the etymon, such as discussions and/or bibliographic references regarding intermediate stages of development, phonetic evolution, concurrent hypotheses, statements of confidence, and secondary etymons.In our system presented herein, we have refined and given more structure to the markup of these datatypes.

*Etymological Link: <etymologicalLink>*

The *etymologicalLink* section is intended to be where the relations between the synchronic and diachronic, or possibly between multiple stages, etymological relationships, or alternative hypotheses of diachronic components are specified and defined. The main way in which this is done in the data structure is through the pointer mechanisms (@source, @target) in the attributes which link the lexical entry to an etymological classification specified by means of an<etymologicalClass> element which can occur within each *etymologicalLink*.

---

[12] Since the paper was written at a time where the XML serialisation of LMF was not yet stabilized, Salmon-Alt (2006) construed an XML representation partially informed with the then ongoing discussions and partially inspired from the TEI Guidelines.

Specified as element values, etymological classes in the model are: inheritance, loan word, word generation, though in the case of disputed word origins, each alternative may have different classifications if need be, and levels of confidence can also be specified (@confidenceScore).

After reviewing the literature on this topic, we can identify several commonalities, the first of which is that all authors looked to the TEI but none found it sufficient to adopt without alterations. Additionally, all works reviewed desire the markup system to have:

- a systematic inventory of typed pointers to link between: etymological forms (etymons) and their synchronic descendants; parallel synchronic forms in related languages (e.g. *cognates*); multiple synchronic forms in a single language; and
- structures dynamic and consistent enough to enable automatic processing, manipulation and evaluation with software applications;
- the ability to classify and assign typological labels to an etymological entry;
- a means of expressing level of certainty of etymological analysis;
- a means of decomposing compounds and components of derivational morphology.

In this paper, we elaborate on the aforementioned sources presented above. We specifically focus on building upon the general model proposed by Salmon-Alt (2006) which is based on a network of etymons and links. Moreover, we identify a more precise group of link types between etymons and explore the consequences in terms of both theoretical implications and possible representations in the TEI framework. Whereas this was not completely stated in Salmon-Alt (2006), we are describing etymological links as the expression of specific etymological processes between etymons (forms), lexical entries or even senses within entries. The rest of the paper is organized to describe the type ontology that we have devised.

# 4. Basic mechanisms for representing etymological processes

## 4.1 An extended TEI-based representation of etymons

The current content model of the <etym> element, as well as the documentation and examples available in the TEI guidelines favor a flat annotation of etymological content that does not put forward whether the actual nature of etymons as references to dictionary entries nor the central role of etymological links in the diachronic processes. Starting from the following example expressed in current recommendations of the TEI guidelines, we show in this section how to go towards a better representation of etymons in etymological description:

```
<entry>
<form type="headword">
<orth>Âbend</orth>
</form>
<gramGrp>
<gen>Mask.</gen>
</gramGrp>
<!-- sense, other info here -->
<etym>
<lang>Ahd.</lang><mentioned>âband</mentioned>,
<lang>mhd.</lang><mentioned>âbent</mentioned>;
<bibl>zur Etym. s. Kluge Mitzka 18. Aufl. unter ,,Abend'', ferner Schwäb. Wb. 1,
11ff.Schweizdt. Wb. 1,34ff.</bibl>
```

```
</etym>
</entry>
```

As we can see, the TEI guidelines have favored so far the use of the <mentioned> element as the basis for marking up etymons.

The first reason why we think this representation is problematic is that it introduces a specific mechanism to refer to lexical items, whereas the TEI dictionary chapter also provides <oRef> and <pRef> in examples and <ref> in external references. Therefore we suggest that such references be considered as a single process of referring to other lexical entries at large, whether within the same entry, the same dictionary, or potentially to a lexical entry from another dictionary. In the latter case, the dictionary may or may not exist for the corresponding language. It may beassumed as a potential construction. This is typically the case for etymons that refer to other languages or ancient forms thereof, even if such forms are not part (yet) of a real lexical description.

To this purpose, we make the recommendation to systematically use <oRef> and <pRef> in all three constructs (examples, etymology and external references), and thus supersede both <mentioned> and <ref> for such usages.

The schematic structure that we propose for this construct is as sketched below:

```
<cit type="etymon">
        <oRef>|<pRef>
        <date>|<gramGrp>|<usg>|<gloss>|<ref>
</cit>
```

Here we see that we can have <oRef> or <pRef> (or possibly both) to refer to the form of the etymon, to which we add further information or constraints related to dating (<date>), grammatical information (<gramGrp>), semantic domain or register (<usg @type>), or translational equivalence (<gloss>). There could of course be additional constraints depending on the complexity of the available etymological information, for instance when an explicit reference to an externally defined sense, beyond the shallow capacity of <gloss>[13], as we shall see later in the paper.

Moreover, the use of <oRef> and <pRef> here is quite important in our model, since it reflects the vision that an etymon is a potential reference to a lexical entry in a dictionary for the corresponding language either synchronically (e.g; in the case of loan words) or, more often, diachronically.

The second flaw with the current TEI proposals for etymology is the lack of mechanisms to group together an etymon with the possible constraints (language, grammar, usage) that may be associated with it. The flat annotation format leaves such pieces of information isolated, as we can see in the previous example for the even central language information (coded with <lang>). Here again, we take up a construct that already exists in the TEI dictionary chapter to encompass this new use case, namely <cit>. By definition, <cit> groups together a linguistic segment with additional features that document its usage and is currently used for examples and translations in dictionary entries. We suggest extending its scope to make it the central construct for the representation of etymons in combination with the use of <oRef> and <pRef> we have just described.

If we take up the preceding example, we can turn it into our suggested representation as follows[14]:

```
<etym>
```

---

[13] A possible replacement for <gloss> could be a construct such as <ref corresp="…" type="sense">

[14] For the sake of conciseness we have not added the bibliographic description (<bibl>) to this example although it would definitely also fit into the <cit> construct outlined here.

```
    …
<cit type="etymon" xml:lang="gmh">
<oRef>âbent</oRef>
<lang>mhd.</lang>
</cit>
    …
</etym>
```
We will show more examples of this construct in the course of our paper, but we can already see how it allows us both to precisely localize etymological descriptions and provide the basic unit for the creation of a generic lexical network across a variety of dictionaries.

## 4.2 Generic representation of an etymological structure

We describe in this section some examples of common types of etymological processes,their linguistic features and key data points, and demonstrations of strategies for encoding each using TEI.

At the most basic level, the origin of any lexical item or sub-form can be described as: a) inheritance from a "parent", "proto-" or predecessor forms of a language; b) borrowing from a foreign language; c) processes that occurred within a contemporary language or sub-varieties of a language[15]. With the exception of a lexical item that was inherited and underwent no change to its surface form (phonetic or phonological), its grammatical role, or its semantic profile, the etymology of a lexical item originating by any one of these means will be comprised of any number of processes that occur on one or more levels of language.

An important aspect to be mentioned at this stage is that etymological description can actually occur at two different levels in the organization of a lexical entry. When one deals with the actual etymology of the word, in the sense of the occurrence of the whole lexical entry in the repertoire of a given language, the <etym> element will obviously appear as child of <entry>. It can also be the case that one has to deal with the emergence of a new sense for a given word, in which case,<etym> should be attached to the corresponding <sense> element. Clearly, this distinction is correlated with etymological types, borrowing being more likely to be related to lexical entries whereas metaphors would rather correspond to new senses.

Any changes that occur within a lexiconcan be labeled in the @type attribute of the <etym> element in a TEI dictionary[16];and the fact that it occurred within the contemporary lexicon (as opposed to its parent language) is indicated by means of @xml:lang on the source form[17].

In the TEI encoding, the former two can be respectively labeled as:
`<etym type="borrowing">…</etym>`
and
`<etym type="inheritance">…</etym>`
Each of the above would represent the top level <etym> element and any other sub-processes can be encoded as embedded <etym> elements with @type attributes.

---

[15] It may of course be the case that the source of a lexical item is unknown.

[16] Currently the use of @type in the <etym> element is not permitted in the TEI schema as the aforementioned element is not a member of the att.typed class. At the time of writing this paper, the proposal has been submitted in the TEI GitHub and is available at the following url (https://github.com/TEIC/TEI/issues/1512). In our proposal, the <etym> element has to be made recursive in order to allow the fine-grained representations we propose in this paper. The corresponding ODD customization, together with reference examples, will be made available on GitHub at the time of publication of this paper.

[17] There may also be cases in which it is unknown whether a given etymological process occurred within the contemporary language or parent system, in such cases the encoder can just use the main language of the entry in both the diachronic <etym> as a default (see for instance example 11).

Alternatively,they can be implicitly encoded as the value of the attribute @xml:langwithin the source <oRef> and/or <pRef>form without having to embed one or more <etym> elements where this information is redundant or implicitly understood, thus simplifying the data structure. For instance in an entry denoting changes to morphological and phonological form of a given Bavarian word inherited from Middle High German, rather than doing the following for every instance of inheritance:

```
<etym @type="inheritance">
<etym @type="phonological-processA">
        …
<pRef @xml:lang="gmh">…</pRef>¹⁸
</etym>
</etym>
```

it may be desirable to simple specify the source etymon form as follows:

```
<etym @type="phonological-processA">
        …
<pRef @xml:lang="gmh">…</pRef>
</etym>
```

As long as the parent or source language(s) of an entry form are known, and declared somewhere project-internal or external ontology or schema, this method allows for a lighter means of expressing the source language of a form, and which can nonetheless still encode implicitly the fact that the origin of the form is borrowed from another language or inherited from a so called parent language.

### 4.3<cit> for etymons and more

Having covered the proposed usage model for<cit type="etymon">we now move on to two further functionsfor the <cit> element, namely to represent (sub-) components of an etymological form (e.g. decomposition) and attestations of their usage.

Components are objects which are important to isolate explicitly within various types of etymologicalformation. They are indeed objects that are very close to etymons but cover a wider variety of linguistic segments such as morphemes for instance. We will see in the course of the paper how a <cit type="component"> makes sense for this purpose.

Attestations of historical forms of an etymon in source context can be included within a citation element as <cit type="attestation">.As the contents of the citation are a quotation, the sampled linguistic content of the attestation is contained within the <quote> element. Within <quote>, the referenced form of the attested etymon can be encoded in the <oRef> element to specify which portion of the text corresponds to the given etymon.

Furthermore, where an attestation of an etymological form is in a language other than the entry itself, it can be necessary to include translations of the attestations, which can simply be encoded with <cit type="translation"> embedded within the <cit type="attestation"> due to the fact that they pertain specifically to the language content of the attestation, they are embedded within the attestation. Thus, the XML structure mirrors that of *attestation*, with the only differences being the value of @type and @xml:lang.

### 4.4 Encoding languages and representational aspects of linguistic forms

Over the course of our research we have constantly been facing issues related to the actual encoding of language related information for headwords in dictionary entries as well as for etymons and similar references. This covers the whole range of TEI elements we are considering here: <orth> and <pron>, with their referential counterparts <oRef> and <pRef>.

---

[18] The language information of an etymon can also be specified within the <cit @type="etymon"> in which the <pRef> or <oRef> is embedded (as explained in section 4.2)

As a basis for our representation, we have of course taken up the use of @xml:lang over these elements together with the constraints applicable to its values and the guidelines of (BCP 47). We will not go into the detail here of the possible limits of this recommendation, instead, we need to elicit some of the choices that are applied in our paper.

First, there is a general issue with the language coverage offered by BCP 47, which is based upon the IANA registry[19], which only offers language tags for a small number of historical languages needed, even from the reduced perspective of Western Europe etymology: Latin "la", Old French "fro", and Middle French "frm"[20].

Additionally, we identify an additional problem withthe abstract context of general language markup BCP 47 recommendations whichspecifies that both the content language and orthographic script be labeled within the @xml:lang attribute. This is neither a conceptually accurate (as an orthographic system is of course not a language), nor a functionally pragmatic means of representing this information.Functionally, there is no difference between orthography and phonetic notation, (with various degrees of nuance and exceptions for logographic, logo syllabic, and other such systems), they are both representations of the language information at some level. With regards to phonetic information, (at least in the dictionary module), the TEI already has a solution for this in @notation which is an obvious necessity in any linguistic data in order to distinguish between the various phonetic transcription systems.

Similarly, we have observed difficulties in relying on the sole @xml:lang attribute to cater for the various ways orthographic forms can actually occur beyond simple script variations: vocalized/non-vocalized in Arabic, kana/kanji in Japanese, competing transliterations systems, etc. Whereas BCP 47 introduces at times ways of dealing with such variations or even allows one to define its private subtags to do so, we found it cumbersome to overload @xml:lang with the main disadvantage of losing systematicity in the way a given language is marked in an encoded text[21]. This is why we have extended the @notation attribute to <orth> in order to allow for better representation of both language identification, and the orthographic content.With this double mechanism, we intend to describe content expressed in the same language.by means of the same language tag, thus allowing more reliable management, access and search procedures over our lexical content.

We are aware that we open a can of worms here, since such an editorial practice could be easily extended to all text elements in the TEI guidelines. We have actually identify several cases in the sole context of lexical representations (e.g. <quote>) here this would be of immediate use.

---

## 4.5 Encoding sequence (diachronic or order of presentation in source)

### 4.5.1 Maintaining Source Structure vs Accurate Representation of Etymological Process

When encoding multi-stage diachronic etymologies fromexisting sources such as attestations from etymological dictionaries, academic papersor otherwise, it may be the case that the source information is not in chronological order for one reason or another. In such cases it is up to the encoder to decide whether maintaining the format of the source is of any benefit to the data quality.

### 4.5.2 Sequence

Because the etymology being encoded in the example has multiple stages in which the sequence is both known and is theoretically relevant, each <cit> should be given an @xml:id in combination with one or both of the sequential pointer attributes: @prev, @next. The combination of these within the data structure encodes relative occurrence of thegiven etymon within the diachrony of any example.

# 5. Inheritance

While *inheritance* isnot itself an etymological process, it identifies lexical items known, or presumed to be inherited from predecessor or 'parent' languages; these forms are sometimes referred to as '*native*'. A simplified view of inheritance is that it is the etymological counterpart to *borrowing*[22]. In the most basic use of an inheritance etymology, an encoder can simply distinguish the given lexical item as having originated directly from its known parent language, or even theoretical proto-language.

However, within the historical trajectory of most inherited lexical items, any number of different etymological processes may occur on every level of language, including phonetic/phonological, phonotactic, morphological, grammatical and/or semantic.

The basic concepts necessary for a minimal encoding of a simple inheritance etymology are:

- language of the (synchronic) entry;
- parent/predecessor language;
- the synchronic orthographic and/or phonetic form(s); sense(s), and/or grammatical information;
- the diachronic orthographic and/or phonetic form(s), sense(s), and/or grammatical information.

The following is a very simple example from Sardinian *semper 'always, still'*, in which the etymology shows that, at least as far as theorthographic form of the lexical item is concerned, it has not changed from its Latin source. This is perhaps noteworthy in its own right, and sampling an entire lexicon in which such information is included could be useful in measuring how much a language has changed over a given period of time.

Example 3: TEI Modeling: <etym type="inheritance">

```
<entry xml:id="semper" xml:lang="srd">
   <form type="lemma">
      <orth>semper</orth>
```

---

[22]The theoretical line between the two (*inheritance* and *borrowing*) becomes blurry when the scope of a language's history is expanded; depending on how far back one looks, an item that was inherited from a direct ancestor, may have been borrowed at an earlier time. Where such cases are known, it is possible to encode both etymological lineages within the same entry.

```xml
        <gramGrp>
            <pos>temporalAdverb</pos>
        </gramGrp>
    </form>
    <sense>
    ...
    </sense>
    <etym type="inheritance">
        <cit type="etymon">
                <oRef xml:lang="la">semper</oRef>
        </cit>
    </etym>
</entry>
```

Note here that we could have an @corresp attribute on <oRef/> if we had a reference Latin dictionary at hand and wanted to point to the actual entry for "semper".

## 5.1 Inheritance & Phonetic/Phonological Changes

For any lexical item regardless of other types of etymological processes undergone over a sufficient span of time, it is of course likely that it will undergo some degree of phonetic changes. Such changes may occur either over a span of time during which a descendant language has become distinct from its 'parent' (*such as Vulgar Latin > French*) or within a span of time in which they are regarded as having occurred within the same language. Phonetic and phonological changes often occur in stages and have their own set of classifications and terminology that require their own level of encoding separate from those occurring on the higher levels of language such as morpho-syntax and semantics.

The basic concepts necessary for a minimal encoding of a phonological etymology within an inherited entry are:

- language of synchronic entry;
- parent/predecessor language at the given stage of etymology;
- the synchronic orthographic and/or phonetic form(s);
- the diachronic orthographic and/or phonetic form(s);
- the relative order of their usage/occurrence (if multiple stages are shown, and their sequence is known);
  Additionally beneficial are:
- dates for each of the diachronic forms;
- bibliographic sources for forms, and for the analysis.

## 5.2 Stages of Phonological Changes in Inherited Forms

This following is our proposal to encode an example of the most significant stages of the phonetic evolution of the French 'chef' from the Vulgar Latin[23] CÁPÚ as per Laborderie & Thomasset (1994). Each <cit type="etymon"> element cluster contains the historical phonetic forms posited by the authors in the <pRef> element, as well as other relevant information pertaining to the given stage in the diachrony of the entry. The etymon clusters begin with the Vulgar Latin form (top) and end with the Middle French form (bottom)[24].

---

[23]Despite the fact that it is widely accepted among researchers that the actual language spoken by most Roman peoples in everyday life was a non-standardized and non-literary language referred to as Vulgar Latin ('VL'), there is no ISO 639 language code for this. Instead, there's just a single tag for Latin (iso 639-3: 'la'). Not distinguishing at least between Classical, Vulgar Latin and Medieval Latin is just not an accurate depiction of the etymological information. Needless to say, this is issue needs to be resolved and one or more proposals for the creation of these tags.

[24]The final form given is the Middle French (and not Modern) because according to the source, it was the final stage in the phonological evolution of that item and is identical to the present day form.

**Example 4: Phonological Stages in Inherited form[25]**

```xml
<entry xml:id="chef" xml:lang="fr">
        <form type="lemma">
                <orth>chef</orth>

                <pron notation="ipa">ʃɛf</pron>
                <gramGrp>
                        <pos>noun</pos>
                        <gen>masc</gen>
                </gramGrp>
        </form>
        <sense>
        ...
        </sense>
        <etym type="inheritance">
                <cit type="etymon" xml:id="kápŭ" next="#kábu">
                        <pRef notation="private" xml:lang="la">kápŭ</pRef>
                </cit>

                <cit type="etymon" xml:id="kábu" prev="#kápŭ"><!-- intervocalic voicing —
>
                        <date notBefore="0350" notAfter="0399"/>
                        <pRefnotation="private"xml:lang="la">kábu</pRef><!--   gallo-latin
or (VL-Gaul) —>
                </cit>

                <cit type="etymon" xml:id="káβǫ" prev="#kábu" next="#t̯áβǫ">
                        <date notBefore="0400" notAfter="0499"/>
                        <pRefnotation="private">káβǫ</pRef><!-- late gallo-latin ?—>
                </cit>

                <cit type="etymon" xml:id="t̯ávǫ" prev="#káβǫ" next="t͡sávǫ">
                        <date notBefore="0400" notAfter="0499"/>
                        <pRefnotation="private">t̯áβo</pRef><!-- late gallo-latin ?—>
                </cit>

                <cit type="etymon" xml:id="t͡sávǫ" prev="#t̯ávǫ" next="#t͡šíęvǫ">
                        <date notBefore="0400" notAfter="0499"/>
                        <pRefnotation="private">t͡sávǫ</pRef>
                        <!-- late gallo-latin ?—>
                </cit>

                <cit type="etymon" xml:id="t͡šíęvǫ" prev="#t͡sávǫ" next="#t͡šíęf">
                        <date notBefore="0450" notAfter="0550"/>
                        <pRefnotation="private">t͡šíęvǫ</pRef>
                        <!-- late gallo-latin ?/early gallo-romance—>
                </cit>

                <cit type="etymon" xml:id="t͡šíęf" prev="#t͡šíęvo" next="#šyęf">
                        <date notBefore="0600" notAfter="0699"/>
                        <pRefnotation="private">t͡šíęf</pRef><!-- early gallo-romance—>
                </cit>

                <cit type="etymon" xml:id="šyęf" prev="#t͡šíęf" next="#šęf">
                        <date notBefore="0700" notAfter="0799"/>
                        <pRefnotation="private">šyęf</pRef><!-- early/Proto Old French (?)
—>
                </cit>

                <cit type="etymon" xml:id="šęf" prev="#šyęf" next="#šęf">
```

---

[25] Whereas in examples, the value of @notation used in <pron> and <pRef> have been well known standard systems with a conventional name; e.g. "ipa" and "xsampa". However in this example, while in the source the author explains the phonetic correlate for the transcription notation used in the work, and there are some characters that are used in the IPA, this system has no proper name, thus we have chosen to label it "private".

```xml
                <date notBefore="1500" notAfter="1650"/>
                <pRef notation="private" xml:lang="frm">šéf</pRef>
        </cit>

        <cit type="etymon" xml:id="šéf" prev="#šéf">
                <date notBefore="1500" notAfter="1650"/>
                <pRef notation="private" xml:lang="frm">šéf</pRef>
        </cit>
        <bibl>Laborderie, N. and Thomasset, C. (1994). Précis de phonétique
historique. Paris: Nathan.</bibl>
        </etym>
</entry>
```

The diachronic sequence of the forms is encoded in our markup as follows: the @xml:id attribute is included for each <cit> for which the given language information is available, the ordering of each is encoded in the data structure by the use of the pointing attributes: @prev and @next, the values of which are the unique identifiers of the previous and next <cit> block respectively.

The <date>[26] element is listed within each etymon block; the values of attributes @notBefore and @notAfter specify the range of time corresponding to the period of time that the given form was is use according to the authors[27]. The attribute values encode the date according to W3C recommendations[28] and must have a four-digit representation of the year[29].

Finally, we have the <pRef> element, which as always, contains the language of the given etymon as the value of @xml:lang and the notation attribute @notation. In this example, the information corresponding to each of the aforementioned represents challenging special cases that is relevant to encoding etymological information as accurately as possible, and in conformance with established standards for best practice in language markup.

## 5.3 Morphological and morphosyntactic changes in inherited forms

Changes in morphological inflection paradigms in which there is no difference need not be explicitly represented in a dictionary but instead can be done implicitly.

Due to the fact that an entry in a TEI or other semasiological dictionary represents the etymology of an individual etymon as pertaining to an individual lexical item, diachronic changes in morphological inflection patterning are manifested in the differences in the phonetic and often orthographic forms. These differences are evident when contrasting a large sample of synchronic and diachronic phonological and phonotactic forms with respect to a given historical or contemporary morphological or morpho-syntactic feature.

When attempting to extract any global information about changes to the grammatical feature inventory of a language from a dictionary, it can also be inferred through the contrast of: the contents of <gramGrp> in the source etymon (in the <cit @type="etymon"> cluster); and the resulting (synchronic) form. For example, where the source form has a specific case and the entry in question does not have any case information, the general phenomena that is the loss of grammatical case is implicitly present.

---

[26] <date> in <cit> is another example which is not adherent to the current TEI standards. We have allowed this within our ODD document. A feature request proposal will be made on the GitHub page and this feature may or may not appear in future versions of the TEI Guidelines.

[27] In the (French language) source of this example (Laborderie & Thomasset, 1994) the dates were given in a combination of roman numerals, superscripted numbers and letters to indicated the century, e.g. IVe2 , which correspond to 'deuxieme moitié du 4e siecle', second half of the 4th century (CE). Optionally, the encoder could include the original date as the value of the <date> element for human readability. In such a case, for the purposes of quality data structuring, compatibility and retrievability the dating information should nonetheless also be included as attribute value(s): e.g. <date notBefore="0350" notAfter="0399"> IVe2 </date>

[28] see http://www.w3.org/TR/NOTE-datetime

[29] One of our anonymous reviewers kindly noted that we should bear in mind that these attributes are for Gregorian dates only. If we are ever encoding anything more specific than a year, it is necessary to use @datingMethod and custom dating attributes to clarify that the Julian calendar is the one we are using here, rather than the proleptic Gregorian.

The following Sicilian entry shows an example of such a scenario in combination with a change in grammatical gender. Whereas in Latin, the etymon had the neuter gender, the inherited form in Sicilian is of the masculine gender, additionally, Sicilian no longer has a neuter gender[30]. Both of these changes are implicit in data structure:

**Example 5: Implicit Morpho-Syntactic Changes: Latin > Sicilian**

```xml
<entry xml:id="mare" xml:lang="scn">
        <form type="lemma">
                <orth>mari</orth>
                <gramGrp>
                        <pos>noun</pos>
                        <gen>masc</gen>
                </gramGrp>
        </form>
        <sense>
        ...
        </sense>
        <etym type="inheritance">
                <cit type="etymon">
                        <oRef xml:lang="la">mare</oRef>
                        <gramGrp>
                                <pos>noun</pos>
                                <gen>neut</gen>
                                <case>nom</case>
                                <iType>-i stem</iType>
                        </gramGrp>
                        <gloss>sea</gloss>
                </cit>
        </etym>
</entry>
```

As with morphology and phonology, grammaticaletymological can be globally inferred. So in a Sicilian TEI dictionary (of sufficient size) containing such synchronic and diachronic information, a search containing the following facets would enable the conclusion that: *there is no more neuter gender*;

For all entries containing:

- <pos>noun</pos>, or <pos>adjective</pos>[31];
- <etym type="inheritance"> in which value of @xml:lang in <oRef> and/or <pRef> is "la");
- <cit type="etymon"> in whose <gramGrp> there is <gen>neut</gen>
- Get synchronic gender: value of <gramGrp>, <gen> within <form type="lemma">[32];

## 5.4 Morphological Inheritance - Inflected Forms

Individual morphologically inflected forms of entries may have different etymological histories than the lemma forms with which they are associated. This could be for reasons connected with a separate origin from the lemma, or more commonly, the fact that the phonological composition of the lemma and inflected forms were and are distinct and they each underwent unique changes.

To encode the etymology any individual form (when there are more than one in the entry that need to be distinguished):

- a unique identifier @xml:id should be included on the given form;

---

[30]This would be referred to as case loss from the perspective of the item's morpho-syntactic etymology.
[31]or any terminological equivalent to either;
[32]which one could easily retrieve by means of a simple XPath instruction such as: ./cit[@type='etymon']/gramGrp[gen = 'neut']

- the corresponding <etym> should point to the form as the value of the attribute @corresp; The following example from Italian shows a case which in addition to phonotactic changes reflected in the orthographic form (e.g. perdidi > persí), the aspect of an inherited inflected form persí has also changed. Specifically, the etymon stemming from what in Latin was the inflection for the past tense, perfective aspect: perdidi, in Standard Italian and many (mainly northern) dialects, descendants of the Latin perfective past have come to be used as the remote past tense (i.e. remote aspect).

**Example 6: Italian - Etymological change in aspect  (perdere)**

```xml
<entry xml:id="perdere" xml:lang="it">
    <form type="lemma">
        <orth>perdere</orth>
        <gramGrp>
            <pos>verb</pos>
        </gramGrp>
        <!-- rest of inflected forms here -->
        <form type="inflected" xml:id="perdere-1s-rem-pt-indic">
            <orth>persi</orth>
            <gramGrp>
                <per>1</per>
                <number>sg</number>
                <tns>past</tns>
                <mood>indic</mood>
                <gram type="aspect">remote</gram>
                <gram type="voice">active</gram>
            </gramGrp>
        </form>
    </form>
    <sense>
        <!-- sense info here -->
    </sense>
    <etym type="inheritance">
        <cit type="etymon">
            <oRef xml:lang="la">perdere</oRef>
        </cit>
    …..
        <etym type="inheritance" corresp="#perdere-1s-rem-pt-indic">
            <cit type="etymon">
                <oRef xml:lang="la">perdidi</oRef>
                <gramGrp>
                    <per>1</per>
                    <number>sg</number>
                    <tns>past</tns>
                    <mood>indic</mood>
<gram type="aspect">perfective</gram>
                    <gram type="voice">active</gram>
                </gramGrp>
            </cit>
        </etym>
    </etym>
</entry>
```

# 6. Borrowing

Borrowing, (*aka., loaning, importing, transferring, copying*) can generally be described as the process in which a lexical item, phrase or other linguistic feature from foreign language or dialect is conventionalized into another language or dialect.

Matras & Sakel (2007) distinguish between two primary types of borrowing, they are *material borrowing,and structural borrowing*. Material borrowing pertains to the borrowing of sound-meaning pairs (e.g. *loanwords, phrases and/or affixes*); structural borrowing pertains to the copying of syntactic, morphological, or semantic patterns.

A major factor in borrowing is contact between a foreign culture and a given language (Haugen, 1950).There are often historical, sociological and practical explanations for how and why the process occurred and thus such information encoded by a linguist or lexicographer may be relevant to those studying other fields such as: history, anthropology, sociology, etc.[33]

Along these lines, distinctions have been made between *cultural borrowings* and *core borrowings* (Myers-Scotton, 2002). Cultural borrowings are when a lexical item is borrowed for a new concept; these can appear over a very short period of time. Core borrowings duplicate existing items and are introduced over a longer period of time by bilinguals (Ibid, Haspelmath, 2003).

When in some manner a borrowed item is not compatible with that of the recipient language, loanwords undergo one or more adaptations to their orthographic, phonological, and/or morphological forms(Haugen, 1950). Our examples will demonstrate encodings for each of these different types of adaptation.

The basic concepts necessary for a minimal encoding of a 'borrowing' etymology are:

- borrowing language (aka. recipient language);
- source language (aka. donor language);
- source form (orthographic and/or phonetic);
- borrowed form (orthographic and/or phonetic);
- semantic profile/meta-linguistic concept or grammatical function of borrowed form (in the case of grammatical borrowing).

The combination of these elements in a TEI based model naturally leads to the overall model depicted in Figure 2.

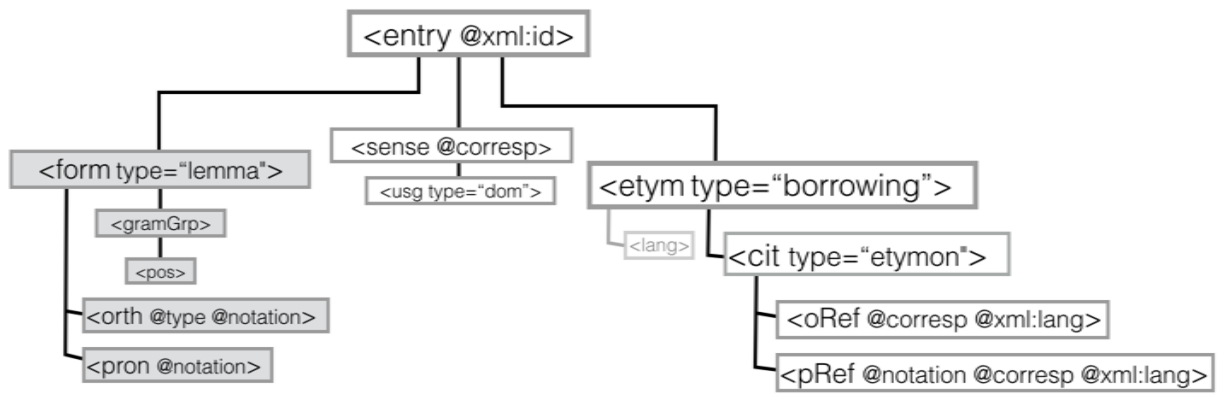

Borrowing-typed etymologies should be direct children of <entry> as the process of borrowing does not affect the sense of a pre-existing lexical item within the target language. The source language should be labeled with the @xml:lang attribute within the <pRef> and/or <oRef>, and can optionally be labeled explicitly for human readability as the value of the element <lang> (*see previous example*). If desired, in addition to the <lang> element, other key terms can be included in the <lbl> element to modify the language (*e.g., <lbl>source</lbl>, <lbl>from</lbl>, etc.,*). If the semantics of the item in the source language differs from that of the borrowing language, the original can be includedin element <gloss>.

---

Example 7 shows our TEI adaptation from LMF (Alt, 2006) of the French 'pamplemousse', which is a borrowing from Dutch. The bibliographic source of this entry is from the *Trésor de la Langue Française* which is cited in the <ref>element both in the value of the attribute @target="#TLF", and within the <ref> element value. The former points to a bibliographic entry for the *Trésor de la Langue Française* that is necessarily located within the given document or project, the latter is included for human readability.

**Example 7: Simple Borrowing 'Pampelmousse' from Salmon-Alt 2006; converted from LMF**

```
<entry xml:id="LE1" xml:lang="fr">
	<form type="lemma">
		<orth>pamplemousse</orth>
		<gramGrp>
			<pos>commonNoun</pos>
			<gen>masculine</gen>
		</gramGrp>
	</form>
	<sense>
	....
	</sense>
	<etym type="borrowing">
		<lang>Dutch</lang>
		<cit type="etymon">
			<oRef xml:lang="nl">pompelmoes</oRef>
			<gloss xml:lang="lat">Citrus maxima</gloss>
			<gramGrp>
				<pos>commonNoun</pos>
				<gen>feminine</gen>
			</gramGrp>
			<note>probablement d'origine tamoule, De Vries, Nedrl</note>
			<ref target="#TLF">TLF</ref>
		</cit>
	</etym>
</entry>
```

## 6.1 Orthographic Adaptation in Borrowed Forms

This following example is a revised version of an entry from the TEI Guidelinesof the word biryani (or biriani) which was borrowed in English from Urdu. Herein we observe that in English, there is variation in the orthographic adaptation of the borrowed item, which is most commonly spelled biryani, but is also sometimes spelled biriani. This relative frequency between the two variant forms is encoded in the entry by embeddingthe less frequent <orth> form within a second <form> element, labeled @type="variant", which is a direct child of the <form type="lemma">.

This orthographic variation stems from the fact that there is not a standard 1 for 1 system of transliterating between the Arabic script of Urdu and Latin script of English. This information is recorded within the <note> element in the etymological section as the origin of this issue is due to the item's etymology.

**Example 8: Borrowing & Transliteration: 'biryani'**

```
<entry xml:id="biryani" xml:lang="en">
	<form type="lemma">
		<form type="preferred">
			<orth>biryani</orth>
			<pron notation="xsampa">%bIrI"A:nI</pron>
		</form>
		<form type="variant">
			<orth>biriani</orth>
		</form>
		<gramGrp>
			<pos>noun</pos>
		</gramGrp>
```

```xml
        </form>
        <sense corresp="http://dbpedia.org/resource/Biryani">
                <def>any variety of Indian dishes…</def>
        </sense>
        <etym type="borrowing">
                <lbl>from</lbl><lang>Urdu</lang>

                <oRef xml:lang="ur">بریانی</oRef>
                <note>The variation in the English orthographic form of this entry is due
to the fact that there is no standard transliteration between English (Latin) and
Urdu's (Arabic) scripts </note>
        </etym>
</entry>
```

## 6.2 Phonological Adaptation in Borrowed Forms

The following is another example in which there is transliteration between the orthographies of the source (English) and borrowing (Japanese) languages. Additionally, Japanese has multiple orthographic systems into which the borrowed form is integrated; each of these systems are identified in the <orth> element as the value of the @notation attribute. In contrast to the previous example (biryani), the presence of a phonetic form in both the source and borrowing languages implicitly shows the difference in the pronunciation between the English source:

**Example 9: Borrowing - Phonological and Orthographical Adaptation: 'takushī'**

```xml
    <entry xml:id="taxi" xml:lang="ja">
        <form type="lemma">
            <orth type="transliterated" notation="rōmaji">takushī </orth>

            <orth notation="katakana">タクシー</orth>
            <pron notation="ipa">takushi:</pron>
            <gramGrp>
                <pos>noun</pos>
            </gramGrp>
        </form>
            <sense corresp="http://dbpedia.org/resource/Taxicab">
                <usg type="dom">Transportation</usg>
            </sense>
            <etym type="borrowing">
                <lbl>source</lbl><lang>English</lang>
                <cit type="etymon" xml:lang="en">

        <oRef corresp="http://en.wiktionary.org/wiki/taxi#English">taxi</oRef>

        <pRef notation="ipa" corresp="http://en.wiktionary.org/wiki/taxi#Pronunciation">
ˈtæksi</pRef>
                </cit>
        </etym>
</entry>
```

## 6.3 Morphological Adaptation in Borrowed Forms

In the case that a language whose grammatical system has a feature such as gender borrows an item from a language that has no grammatical gender, in order to be assimilated into the recipient language, will need to assign it. In the example below showing the French borrowing of the English 'weekend', the lack of gender is implicitly encoded by the absence of a <gen> element within the <gramGrp> element cluster within the <cit type="etymon">[34]. The adaptation made in assimilating this item into the French language is visible in the inverse manner, (e.g., in the inclusion of the <gen>masc</gen>) within the <gramGrp> element cluster within the <form type="lemma">).

---

[34] Alternation to the gender property in the borrowed form is also evident in the first example '*pamplemousse*' as in French it is a masculine noun whereas in Dutch it was feminine.

In modeling this for French, it is especially important to include the gender information because of the implications for the specific forms that will occur in context of and in reference to *weekend* in the context of natural language (e.g. articles, pronouns,determiners, demonstratives, gender forms of adjectives, etc.).

**Example 10: Borrowing - Morphological Adaptation: 'weekend'**

```xml
<entry xml:id="weekend" xml:lang="fr">
	<form type="lemma">
		<orth>week-end</orth>
		<gramGrp>
			<pos>nom</pos>
			<gen>masc</gen>
		</gramGrp>
	</form>
	<sense corresp="http://fr.dbpedia.org/page/Week-end">
		<def>Le week-end (variante weekend, comme en anglais), issu de l'anglais
weekend; ou la fin de semaine est une période hebdomadaire d'un ou deux jours,
généralement le samedi et le dimanche, pendant laquelle la plupart des gens sont au
repos.</def>
	</sense>
	<etym type="borrowing">
		<cit type="etymon">
			<oRef>weekend</oRef>
			<gramGrp>
				<pos>nom</pos>
			</gramGrp>
		</cit>
	</etym>
</entry>
```

# 7. Metaphor

Metaphor is an undeniably one of the primary motors of lexical innovation as well as of (synchronic) polysemy in languages. On the cognitive level, metaphor can be roughly defined as the process of understanding one concept from one domain of experience in terms of another concept from a separate domain of experience. This is mirrored lexically, where one concept in a given domain is described or referred to using lexical item (and/or vocabulary) from another (Lakoff and Johnson, 1980; Lakoff, 1987; Lakoff, 1993).

While this process is not commonly mentioned as a primary category for linguistic change in lexicographical sources such as etymological dictionaries, it is an essential component of linguistic analysis and is relevant to cognitive sciences and anthropology as well.

Metaphorical mappings often follow patterns in directionality; most commonly, the source is a more concrete, or primitive domain and the target is a more abstract, less conceptually primitive domain (Lakoff and Johnson, 1980; Lakoff, 1987; Lakoff, 1993). Examples of what is meant by conceptual primitive domains are: space, temperature, olfaction, location, motion, quantity, body, food, water, etc.

The driving forces for this process are ontological correspondences between the attributes or internal structures of entities in the source domain with those of the target domain (Lakoff, 1993). Where this is true, the potential for mapping is activated, or is rendered salient on the conceptual level, which in turn enables the linguistic encoding of these structures for communicative purposes. Thus, metaphor in language is a surface level reflection or by-product of operations that occur at the cognitive level. Given the fundamental implication of such ideas, not paying attention to metaphor when documenting language change is missing out on an enormous source of incredibly fascinating information.

The basic concepts necessary for a minimal encoding of a 'metaphor' etymology are:

- Source form (the same as target form at time of process),

- Target form (the same as source form at time of process),
- Domain of concept (y): Source Domain,
- Domain of concept (x): Target Domain,

The model below shows an abstract representation of the essential structures for such an entry.

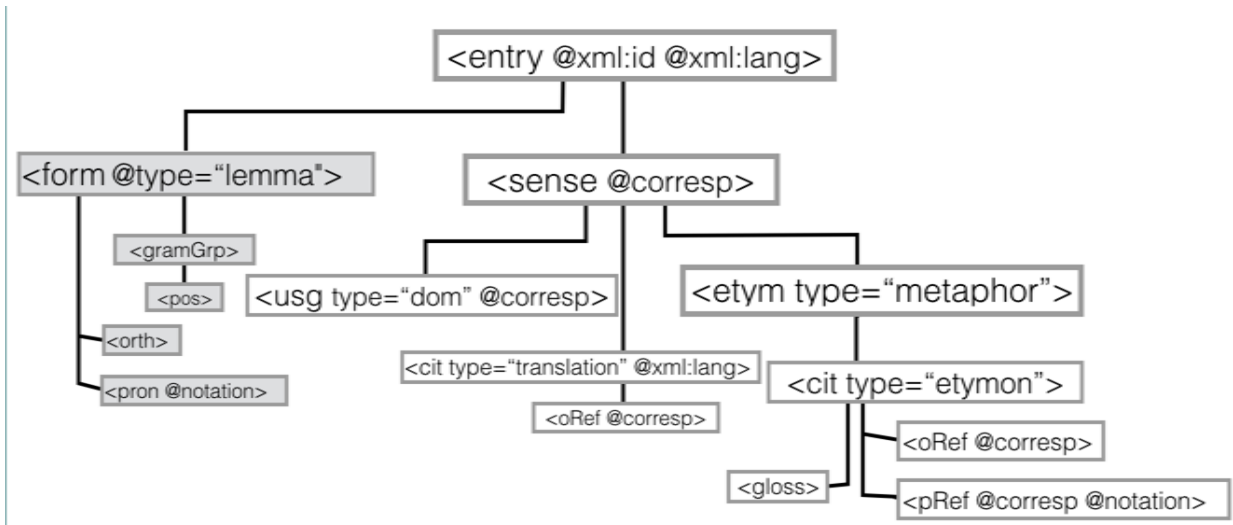

Example 11 shows an instance of an encoding of the lexical item for '*kidney*' in Mixtepec-Mixtec language, which was derived metaphorically from '*bean*'. This metaphorical mapping is clearly due to the physical similarity (color and shape) between a bean and a kidney. While both terms refer to concrete (physical) concepts, the source, which is a primary food source for the speakers of the language, is more conceptually primitive than its target, which is an internal organ and thus seen much less frequently by most speaker (if at all).

Example 11: Metaphor: Mixtepec-Mixtec 'kidney'[35]

```
<entry xml:id="kidney" xml:lang="mix">
        <form type="lemma">
                <orth>ntuchí</orth>
                <pron notation="ipa">ndú.t͡ʃì</pron>
                <gramGrp>
                        <pos>noun</pos>
                </gramGrp>
        </form>
        <sense corresp="http://dbpedia.org/resource/Kidney">
                <usg type="dom"
                        corresp="http://dbpedia.org/resource/Human_body">Body</usg>
                <usg type="dom"

        corresp="http://dbpedia.org/resource/Human_organs">InternalOrgans</usg>
                <etym type="metaphor">
                        <cit type="etymon">
                                <oRef corresp="#bean">ntuchi</oRef>
                                <pRef notation="ipa" corresp="#bean">ndú.t͡ʃì</pRef>
                                <ref                                        type="sense"
corresp="http://dbpedia.org/resource/Bean"/>
                                        <usg                                type="dom"
        corresp="http://dbpedia.org/resource/Category:Edible_legumes">Legume</usg>
                                <gloss>bean</gloss>
                        </cit>
```

---

[35]In the digital source quoted here, entries are associated with an identifier (xml:id) named from the English gloss. Besides, different senses correspond to different entries. #bean thus refer to the entry for 'ntuchi' with the meaning 'bean'. In many cases, such a reference would normally point to senses instead of separate entries.

```
            </etym>
            <cit type="translation" xml:lang="en">
                <oRef>kidney</oRef>
            </cit>
        </sense>
</entry>
```

Metaphor-typed etymologies should always be direct children of <sense> as the lexical change affects the sense of an existing lexical item. In order to know that it is a metaphor, it is of course necessary to know a) the original meaning of the source (encoded in <gloss>, or <def>), and b)the domains of the source and target items, e.g. source domain (*semantic domain of etymon*) & target domain (*semantic domain of entry*). In pursuit of the latter, the TEI <usg> element can be used with the attribute value pair: @type="dom", in which the element value represents the domain. Given that the semantic profiles of many concepts are not always limited to a single domain, it may be necessary to include multiple <usg type="dom"> element-value pairs as is the case in the example. In our model, <usg type="dom"> is used both in the synchronic (<sense>) and diachronic (<cit>) portions of the entry.

However, although such structures, if used consistently, do have the capacity to greatly enhance any data that doesnot include this information, they inevitably fall short of the extent of knowledge that is truly necessary to create an accurate model of metaphorical processes. To do this it is necessary to make use ofone or more ontologies, which could be locally defined within a project, external linked open datasources such as dbpedia, wikidata, or some combination thereof. Within TEI dictionary markup URI's for existing ontological entries can be referenced in the <sense>, <usg>, and <ref> elements as the value of the attribute @corresp.

Within the etymon, the <oRef> and/or <pRef> can be included with a pointer to the source form using the attribute value pair @corresp and the value of which being a reference to the source entry's unique identifier (if such an entry exists within the dataset). In such cases, the etymon pointing to the source entry can be assumed to inherit the source's domain and sense information, and this information can be automatically extracted with a fairly simple XSLT program, thus the encoders may choose to leave some or all of this information out of the etymon section. However, in the case that the dataset doesnot actually have entries for the source terms, or the encoder wants to be explicit in all aspects of the etymology, as mentioned above, the source domain and the ontologically based sense of an etymon can be encoded within <cit> as <ref>and <usg> respectively.

## 8. Metonymy

Like metaphor, metonymy is another process based in human cognition which is reflected in language and leads to synchronic lexical polysemy. Again like metaphor, metonymy is often central to linguistic etymological analyses but is also not commonly mentioned as a primary category for language change in lexicographical sources such as etymological dictionaries.

In metonymy, one vehicle entity provides mental access to a target entity within a single domain by highlighting different aspects of part-whole relationships (meronymy) (Kövecses & Radden, 1998). There are two primary types of relationships that commonly provide the motivation, or what Vandeloise (1999) refers to as 'Logical Impetus' for the occurrence of metonymy, these are: a) where a part of an entity's profile stands for the whole; b) when a domain's sub-part stands for its whole (ibid).

The basic concepts necessary for a minimal encoding of a 'metonymy' etymology are:
- synchronic (polysemous) form(s) of item;
- source form(s) (orthographic and/or phonetic);

- semantic profile/meta-linguistic concept or entity of borrowed form[36];

The TEI modeling of metonymy is very similar to that for metaphor with the exception that the semantic domain between the source and target is the same. Thus the model is as follows: metonymy-typed etymologies should be embedded in the <sense> element, which can also contain the @corresp with a URI to the conceptual entry in an external ontology; within the <cit type="etymon"> element block, the source orthographic and/or phonetic forms should of course be included in <oRef> and/or <pRef> elements; within these elements, the value of the attribute @corresp identifies the unique identifier of the source entry. Often metonymy leads to a change in grammatical role and thus it is often the case that a <gramGrp> element block is necessary, which should of course include whatever sub-elements necessary.

**Figure 4: Abstract TEI modeling for metonymy**

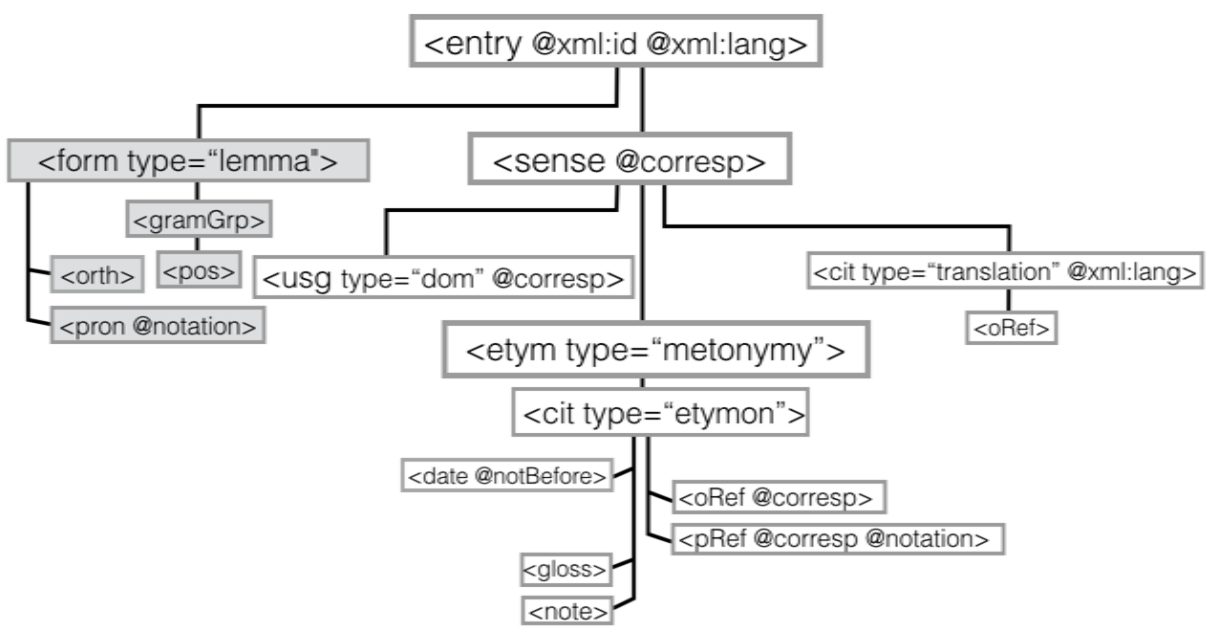

In the example below from Mixtepec-Mixtec, the words for 'animal' and 'horse' are polysemous, the origin of which can clearly be identified as metonymy. Here, the language reflects the history of the people; since there were no horses in Mexico until the arrival of the Spanish, there was no Mixtecan word for 'horse', thus they categorical noun for 'animal' was used to describe the unnamed animal.

In the TEI entry, the synchronic orthographic and phonetic forms are identical to those of the source term 'animal'. Within the etymology element block (<etym type="metonymy">) and the etymon (<cit type="etymon">) the source term's URI is referenced in <oRef> and <pRef> as the value of @corresp (@corresp="#animal").

In <sense>, the URI corresponding to the dbpedia entry for 'Horse' is the value for the attribute @corresp. Additionally, the <date @notBefore> element-attribute pairing are used to specify that the term has only been used for the 'horse' since 1517 at maximum *(corresponding* *of course to the first Spanish expedition into Mexico).* Within the actual document, the contents of <note> discuss the historical context of this word origin, (though this has been commented out in the example as it has already been discussed above).

```xml
<entry xml:id="animal-horse" xml:lang="mix">
    <form type="lemma">
        <orth>kiti</orth>
        <pron notation="ipa">kì.ʈí</pron>
        <gramGrp>
            <pos>noun</pos>
        </gramGrp>
    </form>
    <sense corresp="http://dbpedia.org/resource/Horse">
    <usg type="dom" corresp="http://dbpedia.org/resource/Animal">Animal</usg>
    <etym type="metonymy">
        <date notBefore="1517"/>
        <cit type="etymon">
            <oRef corresp="#animal">kiti</oRef>
            <pRef notation="ipa" corresp="#animal">kì.ʈí</pRef>
            <gloss>animal</gloss>
        </cit>
        <note><!-- notes on historical context of term here --></note>
    </etym>
    <cit type="translation" xml:lang="en">
        <oRef>horse</oRef>
    </cit>
    <cit type="translation" xml:lang="es">
        <oRef>caballo</oRef>
    </cit>
    ….
    </sense>
</entry>
```

# 9. Compounding

For various reasons, the lexical process of compounding has traditionally been treated in linguistic literature focused on synchronic themes, often in the context of such issues as morphology and word formation. One of the least theoretically controversial reasons for this is that novel compounds can be formed, used and understood by speakers instantaneously. However, there is significant discord in the linguistics literature as to how compound lexical items and the process of compounding should be analysed. This is most distinctly evident in the contrasting approaches between linguistic theories that make strict divisions between the lexicon and grammar (traditional generative grammar) and those that do not make such distinctions (cognitive linguistic theories) (Langacker, 1987, 1991, 2000; Jackendoff, 2009).

In attempting to create typologies and/or generalizations about the nature of compound word forms and the process in which they are created, the aforementioned viewpoints emphasize different aspects of the phenomena. Studies by authors adhering to more formal, generative theoretical frameworks often focus on analyzing the morphosyntactic makeup of compound forms in the context of rule patterning: e.g. N+N, A+A, V+V, N+A, P+N, etc. Those adhering to the cognitive based theory generally focus on the semantic profiles and ontological relationshipsof the constituent parts, as well as the cognitive processes (metaphor and/or metonymy) active in the resulting meaning (Goossens, 1994; Geeraerts, 2002; Benczes, 2006a,b; Guevara & Scalise, 2008). Examples of such semantic/ontological relationships commonly found in compound forms are: type-of (hyponymy), cause-effect, part-whole (meronymy), location-located, etc. (Benczes, 2006b).

A common means of classifying compounds in linguistics literature (both formal and cognitive) is according to two main typologies: *endocentric* and *exocentric* compounds (Bloomfield,1933)[37]. There are numerous different typological interpretations according to which the concepts of *endocentric* and *exocentric* compounds are defined in the literature. Many of these differences are attributable to problems with the classification criteria, the types of compounds examined, and the languages analyzed (Guevara & Scalise, 2005; Benczes, 2006a).

Generally, endocentric compound constructions have a '*head*' element which is the hypernym while the full compound represents a sub-classification or hyponym of that head (Benczes, 2006a,b). An example of this is the bird *scrub jay* the head of which is *jay* and the compound *scrub jay'* is a *type of* jay. Exocentric compounds are considered '*headless',* and while there is no single standard ontological relationship between the compound and/or its components; they can contain any variety of relations other than hyponymy[38]. An example of this from Benczes (2006b) is the English compound *hay fever,* which has a cause-effect relation as *hay* causes the effect of *fever*.

While it would be impossible to give a full account of all of the various typologies proposed in the literature,our markup framework for compounding is capable of handling the key pieces of information for a wide array of theoretical approaches for explicit and implicit encoding of the relevant information within a single entry.

Basic pieces of information for modeling compounds and compounding include:
- the delimitation between the components;
- the grammatical profiles of the components;
- the meanings (sense) of the components;
- the semantic domain of the components

## 9.1 EncodingCompounds: Synchronic Portion

Modeling compounding in TEI dictionaries can (and ideally) should be done at both the synchronic and diachronic portion of an entry. On the synchronic level, the attribute-value @type="compound" should always be placed in the <entry> element in order to specify that the contents of form (<orth> and or <pron>) can be further parsed.

Example 13: Compounding (synchronic encoding); French 'merle noir'

```
<entry xml:id="merle-noire" type="compound"xml:lang="fr">
        <form type="lemma">
                <orth>merlenoir</orth>
                <gramGrp>
                        <pos>nom</pos>
                        <gen>masc</gen>
                </gramGrp>
        </form>
                ….
</entry>
```

From a processing point of view, this is of particular use when the compound form includes whitespace. Optionally, the editor may choose the more thorough method of encoding the individual component strings of a compound form with <seg>. Where corresponding entriesexist within a project for each of these portions of the compound, an editor may choose to connect

---

them with the components encased within <seg> using @corresp with the value being the unique identifier for the given entry.

The following example shows the encoding of the number thirteen in Mixtepec-Mixtec which is a case of an exocentric compound as there is no 'head'. As is the case with the lexical composition of numbers in many languages, this compound form is the sum of its component parts semantically and phonologically; e.g. utsi 'ten' + uni 'three'.

**Example 14: Exocentric Compounding (synchronic encoding); Mixtepec-Mixtec 'utsi uni'**

```
<entry xml:id="num-13" type="compound"subtype="exocentric"xml:lang="mix">
        <form type="lemma">
                <orth>
                        <seg corresp="#num-10">utsi</seg><seg corresp="#num-3">uni</seg>
                </orth>
                <gramGrp>
                        <pos>cardinalNumber</pos>
                </gramGrp>
        </form>

        <!-- sense, etym, translation,etc... -->
</entry>
```

Where there is a hyphen in a compound, this can be differentiated from the content portion of the string by using the TEI character element <pc>. The French example below is a type of bird whose orthography is usually spelled with a hyphen but which is also occasionally written as a continuous word without whitespace or a hyphen:

**Example 15: Compounding (synchronic encoding); French 'rouge-gorge','rougegorge'**

```
<entry xml:id="rouge-gorge" type="compound"xml:lang="fr">
        <form type="lemma">
                <orth>
                        <seg>rouge</seg><pc>-</pc><seg>gorge</seg>
                </orth>
                <form type="variant">
                        <orth>
                                <seg>rouge</seg><pc></pc><seg>gorge</seg>
                        </orth>
                </form>
                <gramGrp>
                        <pos>nom</pos>
                        <gen>masc</gen>
                </gramGrp>
        </form>
                ...
</entry>
```

## 9.2 EncodingCompounds: Diachronic Portion

Below is the diachronic portion of the entry of the Mixtepec-Mixtec number 'thirteen' discussed above.

**Example 16: Compounding; Mixtec-Mixtec 'utsi uni'**

```
<entry xml:id="num-13" type="compound"xml:lang="mix">
        <!-- form, gramGrp, etc..-->
        <sensecorresp="http://dbpedia.org/resource/10_(number)">

        <usg type="dom" corresp="http://dbpedia.org/resource/Category:Cardinal_numbers">
CardinalNumbers</usg>
        </sense>

        <etym type="compounding">
                <cit type="etymon">
                        <oRef corresp="#num-10">utsi</oRef>
                        <gramGrp>
                                <pos>cardinalNumber</pos>
```

```
                </gramGrp>
                <gloss>ten</gloss>
        </cit>

        <cit type="etymon">
                <oRef corresp="#num-3">uni</oRef>
                <gramGrp>
                        <pos>cardinalNumber</pos>
                </gramGrp>
                <gloss>three</gloss>
        </cit>
    </etym>
        ….
</entry>
```

The components of the compound which were encased in the <seg> element within the synchronic forms correspond to the etymons on the diachronic level; as in any other etymological process in our encoding system, each sub-unit of the compound and its given lexical information is represented as a separate <cit type="etymon">. Each of these citation-etymon crystals is embedded within the compound-typed <etym>. Since compounds do not represent new senses of a *single* existing entry form, the etymology element is placed as a direct child of <entry>. As we have seen before the attribute @corresp in the <oRef> and/or <pRef> points to the URI for entries of each component of the compound; in this example these entries are within the same document. The comprehensive model is depicted in figure 5 below.

**Figure 5: TEI Modeling Diagram: compounding(Diachronic Portion)**

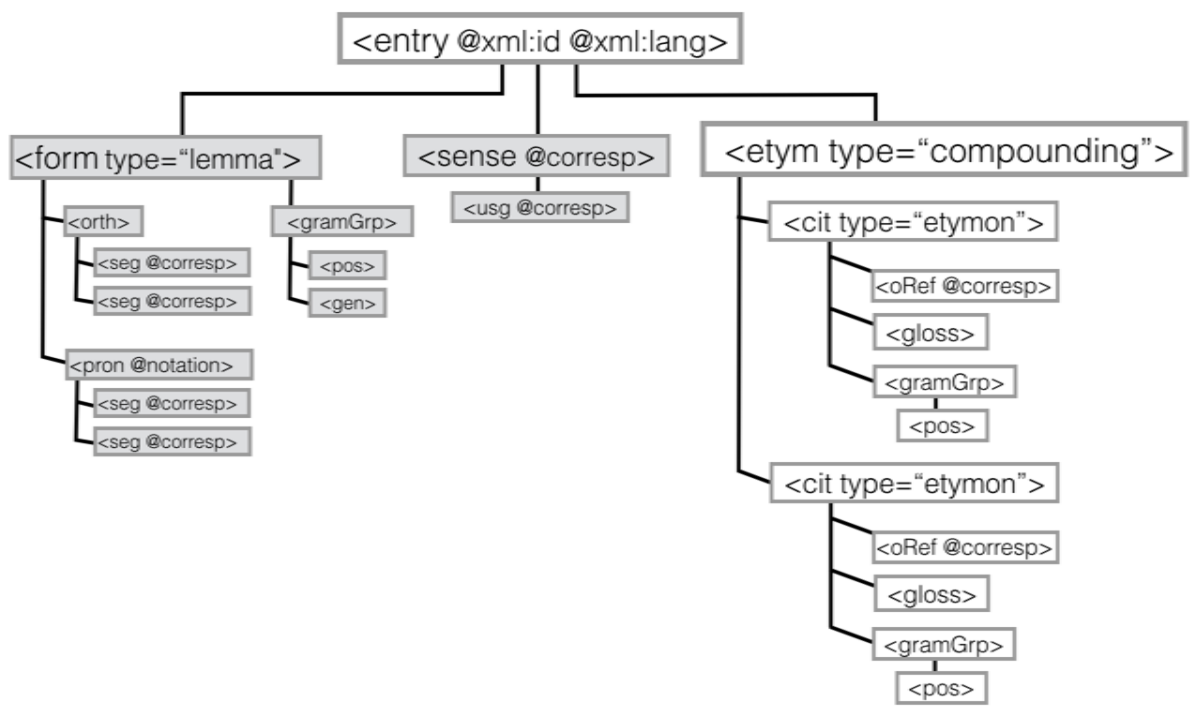

## 9.3 Co-occurrence of Compounding & Other Etymological Processes

### 9.3.1 Compounding and Metaphor

The German item *handshuh* '*gloves*' is an example of a compound in which metaphor is clearly present; shuh *'shoe',* which is an article of clothing specific to the feet (e.g. domain = FEET), is used as an article of clothing for the hands, (e.g. domain = HAND). In the encoding of this item in the entry below, the structure is a combination of everything we have previously seen in examples of compounding- and metaphor-typed etymologies, but with the difference that since <etym type="metaphor"> is a subpart in the larger word formation process of compounding, it is embedded within the <etym type="compound">. Since *compounding* is the top level component of the etymology, <etym> is placed as a direct child node of <entry>, following the guidelines for that element.

Also noteworthy is that this example is an 'endocentric' compound as the *head* is 'shue' '*shoe*'.[39]

**Example 17: Compounding & Metaphor[40] - German 'handshuh'**

```xml
<entry xml:lang="de" xml:id="handschuh" type="compound"subtype="endocentric">
    <form type="lemma">
        <orth><seg>Hand</seg><seg>schuh</seg></orth>
        <pron notation="ipa"><seg>ˈhant</seg><seg>ʃuː</seg></pron>
        <gramGrp>
            <pos>subst.</pos>
            <gen>mask.</gen>
        </gramGrp>
    </form>
    <sense corresp="http://dbpedia.org/resource/Handschuh">
        ...
    </sense>
    <etym type="compounding">
        <cit type="etymon">
            <oRef xml:lang="de">Hand</oRef>
            <pRef xml:lang="de" notation="ipa">ˈhant</pRef>
            <gloss>hand</gloss>
        </cit>
        <etym type="metaphor">
            <cit type="etymon">
                <oRef xml:lang="de">Schuh</oRef>
                <pRef xml:lang="de" notation="ipa">ʃuː</pRef>
                <gloss>shoe</gloss>
            </cit>
        </etym>
    </etym>
</entry>
```

The classification of a lexical item as a compound hinges on the semantic transparency of its components by the speakers of a language. An item corresponding to a single conceptual entity that was derived through compounding in a source can be borrowed by another language and the fact that it was a compound is not relevant to the meaning in the target lexicon. In such cases it is up to the encoder whether or not to include the etymological details that occurred in source language. Below is the full encoding of the earlier example of 'pamplemousse' from Salmon-Alt (2006).

**Example 18: Borrowing & Compounding - 'pamplemousse Salmon-Alt (2006)**

```xml
<entry xml:id="LE1" xml:lang="fr">
    <form type="lemma">
        <orth>pamplemousse</orth>
        <gramGrp>
            <pos>commonNoun</pos>
        </gramGrp>
```

---

[39]While not shown here, if editor desire to label the *head* of the compound it could be done using the attribute @ana the value of which can correspond to a feature structure defined and declared within the project. Where tagging within the synchronic portion of the entry, @ana should be included in the <seg> of the head component. Where tagging within the diachronic portion, @ana should be included on the <cit type="etymon"> of the head component of the compound. The ISOcat has an entry for the category '*head*' whose PID is: http://www.isocat.org/datcat/DC-2306
[40]Let it be noted that the structure for modeling: *compounding + metaphor* applies in the exact same manner to *compounding + metonymy* which is not shown in these examples.

```
        </form>
        <sense>
        ....
        </sense>
        <etym type="borrowing">
                <lbl>source</lbl><lang>Dutch</lang>
                <cit type="etymon" xml:id="L2">
                        <oRef xml:lang="nl">pompelmoes</oRef>
                        <gloss xml:lang="lat">Citrus maxima</gloss>
                        <gramGrp>
                                <pos>commonNoun</pos>
                                <gen>feminine</gen>
                        </gramGrp>
                        <note>probablement d'origine tamoule, De Vries, Nedrl</note>
                        <ref target="TLF">TLF</ref>
                </cit>
                <etym type="compounding">
                        <cit type="etymon">
                                <oRef xml:lang="nl">pompel</oRef>
                                <gramGrp>
                                        <pos>adjective</pos>
                                </gramGrp>
                                <gloss>gros, enflé</gloss>
                        </cit>
                        <cit type="etymon">
                                <oRef xml:lang="nl">limoes</oRef>
                                <gramGrp>
                                        <pos>commonNoun</pos>
                                </gramGrp>
                                <gloss>citron</gloss>
                        </cit>
                        <ref target="#Boulan-König">Boulan, König...</ref>
                </etym>
        </etym>
</entry>
```

### 9.3.2 Lexicalization

Compounds, like morphologically derived coalesced forms and idiomatic forms or other types of phrases, may or may not become integrated into the lexicon of a language. When they do, the process may span over an extended period of time, during which the original meaning ofpart or all of the components may become oblique to speakers[41] to the degree that the item is not recognized as the sum of its parts; this phenomena is called *lexicalization* (Jackendoff, 2009). While lexicalization is indeed a lexico-cognitive process with etymological implications, it is not a process that can be summarized and represented in a dictionary, as there are often different degrees of lexicalization between different speakers.

# 10 Grammaticalization

A major source of change on multiple levels of language is due to the process of grammaticalization in which a lexical item (or a sense of the item) corresponding to a more concrete concept undergoes changes to its semantic profile and can in certain linguistic contexts become more grammatical in nature. Due to the complexity of the process and the continuing debate as to thedetails of just what it entails, any single definition of grammaticalization inevitably raisestheoretical debate, and may potentially be modified down the line within the field. Antoine Meillet (1912), who coined the term, described it as follows:

*"The development of grammatical forms by progressive deterioration of previously autonomous words is made possible by...a weakening of the pronunciation, of the concrete sense of the words,*

---

[41]An example of this in English is horseradish, which speakers can parse the components of *horse* and *radish* however the antiquated figurative sense of *horse* which meant "strong, large, coarse"which gave rise to the compound is no longer used in modern English (source: http://www.etymonline.com/index.php).

*and of the expressive value of words and groupings of words. The ancillary word can end up as an element lacking independent meaning as such, linked to a principal word to mark its grammatical role."*

Studies of grammaticalization have shown that there are certain tendencies for forms to undergo the same types of functional shifts to their semantic, syntactic, morphological and phonological properties, which occur in relatively similar orders[42](Svorou, 1994; Croft, 2003; Hopper & Traugott, 2003). These evolutionary pathways are commonly referred to as 'clines' (ibid). On the semantic level, the most generic characterization of the tendency is for the profile to undergo changes from concrete to abstract (ibid). Moreover there is significant evidence that the semantic profiles of a lexical sources may motivate/constrain the potential structural characteristics and grammatical roles that emerge from a lexical source (Svorou, 1994; Hopper and Traugott, 2003).

Parallel to the semantic shifts are the clines that occur on the syntactic and morphological levels of which there are a number of different specific possibilities, but which can be generalized as follows:

less grammatical → more grammatical

Hopper and Traugott (2003) discuss a correlation between the degree of grammaticalization (grammatical status) and the loss of certain morphological and syntactic properties that are commonly observed in prototypical members of major grammatical categories (*Decategorization*).

majority category > (intermediate category) > minor category

It should be noted that due to polysemy,a single form may simultaneously have several grammatical functions in which multiple 'stages' in different clines can co-exist synchronically (Svorou, 1994; Hopper & Traugott, 2003). According to Svorou (1994), the study of grammaticalization requires a "*panchronic*" view of language change, in whichthe component evolutionary processes are seen as continuous, rather than artificially discrete concepts of synchronic and diachronic[43].

The negotiation of meaning by speakers, through the use of metaphor and metonymy to extend the contexts in which a lexical item is used, enable such changes as pragmatic/semantic bleaching or enrichment on the cognitive, semantic and pragmatic levels. On the syntactic and morphological levels, these higher-level changes are manifested in *reanalysis*[44]and *analogy*[45].

Grammaticalization is thus very complex and involves multiple possible sub-processes which may or may not co-occur on different levels, so encoding such etymologies may require the use of any number of the strategies discussed above.

## 10.1 Overview of Basic Encoding Recommendations

Since grammaticalization potentially involves multiple etymological processes on any level of language at different stages in the diachrony of an item, the formatting recommendations are dependanton the specifics of a given scenario and involve the integration of concepts and data structures from multiple processes discussed elsewhere in this paper. It is up to the editor to determine what aspects and features of the etymology they want to emphasize within an encoding of grammaticalization (eg. processes on different levels, bibliographic attestations, discourse, semantics, morphosyntax, phonology, etc.); its nuances and complexities make it impossible to write an entirely comprehensive set of encoding recommendations. However, the constants which can always be included in a grammaticalization etymology are as follows:

- given that grammaticalization inherently entails a shift in the sense of an etymon <etym type="grammaticalization">should be embedded within <sense>;
- all forms of the etymon that are hypothesized or attested as having occurred within the span of the time corresponding to the stages of the grammaticalization process should be <cit type="etymon">*(as we have previously seen)*:

---

[42]Examples of some of the most common grammaticalization patterns (eg. 'clines') include the following (from Croft 2003): noun > adposition; adposition > case affix; verb > classifier; demonstrative or article > gender/ noun class marker; main verb > tense, aspect, mood marker.
[43]Svorou's panchronic view of grammaticalization is supported by Hopper and Traugott (2003) who state that clines are both diachronic (schema of evolution) and synchronic (all/multiple forms of evolutionary stages may co-exist).
[44] Reanalysis is defined by Hopper and Traugott (2003) as a linear, syntagmatic, reorganization and "rule" changes and is not directly observable.
[45]Analogy involves paradigmatic organization, changes in surface collocations and patterns of usage, which in turn make unobservable changes from reanalysis observable (Hopper and Traugott, 2003).

Conceptual information necessary in the encoding of grammaticalization includes:

- the synchronic form(s)[46];
- the grammatical information;

If the grammaticalized item is one of multiple senses of a polysemous form[47]:

- the grammatical, semantic and/or pragmatic context in which the given sense occurs;
- the collocates with which it occurs (if any);
- the source form(s);
- whether the source form(s) is/are: *attested or hypothesized*;

If attested:

- bibliographic source of attestation;
- date/date range for each attested form;

If hypothesized:

- the semantic profile of source: (e.g. semantic domain; metalinguistic concept) ;
- the grammatical profile of the source;

The placement of the <etym type="grammaticalization"> is dependent on the specific diachrony of the etymon, for example, if the item was inherited, it can be placed within a <etym type="inheritance">.

## 10.2 Traugott (1995): Grammaticalization of English discourse marker *'besides'*

In the following example from Traugott (1995), in our encoding, we show a sample of several stages of the grammaticalization process of the English discourse particle 'besides'. The author makes the case that the following cline (e.g. the process of grammaticalization for a given lexical item) should be added to the scope of grammaticalization phenomena:

<div align="center">(Adverbial of Extension > Sentence Adverbial > Discourse Particle[48])</div>

The major stages of this process are presented in a semi-chronological order and in terms of the following stages:

- stage 0: Full lexical noun;

---

[46]Given the prominent role of phonological change in grammaticalization (e.g. coalescence, phonological reduction, loss), inclusion of <pron> is recommended though it of course is not always possible to present such forms with full certainty (in which case there is always the TEI attribute @cert).

[47]While not 100% essential, including any known contextual, pragmatic, or collocational information greatly enhances the usefulness of the synchronic data modeling and it enhances the precision of the etymological account.

[48] In Traugott (1995), Discourse Markers are a subtype of Discourse Particle.

- stage I: Adverbial of Extension;
- stage II: Sentence Adverbial;
- stage III: Discourse Marker/Particle[49]

The example below is an encodedrepresentation of the arguments put forth by the source and we have attempted to represent the linguistic information, terminology and the etymological processes as accurately as possible without adding or subtracting from the original based on our own interpretation or external data.

**Stage 0:** Full lexical noun

```xml
<sense>
    ....
    <etym type="grammaticalization">

    <cit type="etymon" xml:id="at-850-950" next="#at-1225-a">
        <date notBefore="0850" notAfter="0950"/>
        <oRef xml:lang="ang">
            <seg xml:id="e1s1">sid</seg>
            <seg xml:id="e1s2">an</seg>
    </oRef>
        <gloss>side</gloss>

        <gramGrp><!-- inherently applies to the top level structure -->
            <pos>locativeNoun</pos>
    </gramGrp>

        <cit type="component" corresp="#e1s1">
        <gramGrp>
                <pos>noun</pos>

        </gramGrp>
            </cit>[50]

        <cit type="component" corresp="#e1s2">
        <gramGrp>
                <case>dative</case>

        </gramGrp>
    </cit>
        <cit type="attestation" xml:lang="ang">
            <quote>& þonne  licge  on  ða  swiðran  <oRef>sidan</oRef> gode
hwile</quote>
            <cit type="translation" xml:lang="en">
                <quote>and  then  lie  on  the  right  side  for  a  good
while</quote>
            </cit>
    </cit>
        <bibl>(850-950 Lacnunga Magicand&Med., p. 120 [HC])</bibl>
</cit>
```

This first stage '*sidan*' is from Old English (iso 639-3: 'ang'); as usual the <cit> element is typed as @type="etymon".

- noun inflected with the dative case, both labeled in their respective elements within <gramGrp>and is labeled as <pos>locativeAdverbial</pos>;

The source of this is dated sometime between 850-950. As we have discussed previously, a range of dates can be expressed using the attributes @notBefore and @notAfter in the <date> element. While the bibliographic source (the last direct child element of <cit type="etymon">) does contain the date itself, using the <date> element helps keep the consistency of the data structure which will likely have benefits to any automatic retrieval processes.

---

[49] Note that in the encoding "discourse markers" are represented as the (non-native TEI) element <particle> with the value of 'discourseMarker'. The concept of a lexical '*particle*' is well established in linguistics and lexicography. This element is defined in our ODD and is linked to the ISOcat uri in the declaration.

[50] Additional information related to this etymon could be added here, for instance that 'sid' is a stem and the head of the inflected noun form. This would be a typical application of @ana on the <cit> element.

- a partial example of the context in which the etymon form was observed is given in the <quote>element whose parent is an embedded <cit @type="attestation">. Also specified here is the language attribute and tag, which as alwaysare expressed as: @xml:lang="ang". The form itself is encased in a <oRef> with no attributes.
- finally, embedded within the attestation is the translation of the Old English source sentence into Modern English; e.g. <cit @type="translation" xml:lang="en">

These next two attested etymons are both from 1225 (*Middle English*), and from the same bibliographic source, yet they have different grammatical functions; by including the range of dates for each with their contrasting grammatical usage, the encoded information is able to capture the emerging state of polysemy for the lexical item.

- each have the prefix *bi-*
- each is dative case;

```
<cit type="etymon" xml:id="at-1225-a" prev="#at-850-950" next="#at-1225-b">
    <oRef xml:lang="ang">
        <seg xml:id="e2s1">bi</seg><pc>-</pc>
        <seg xml:id="e2s2">sid</seg>
        <seg xml:id="e2s3">en</seg>
    </oRef>
    <gloss>at his side</gloss>
        <gramGrp>
            <pos>locativeAdverbial</pos>
        </gramGrp>
        <cit type="component" corresp="#e2s1">⁵¹
            <gramGrp>
                <pos>adverb</pos>
            </gramGrp>
        </cit>
        <cit type="component" corresp="#e2s2">
            <gramGrp>
                <pos>noun</pos>
            </gramGrp>
        </cit>
        <cit type="component" corresp="#e2s3">
            <gramGrp>
                <case>dative</case>
            </gramGrp>
        </cit>
        <cit type="attestation" xml:lang="ang">
            <quote>His pic he heold <oRef>bi-siden</oRef></quote>
            <cit type="translation" xml:lang="en">
                <quote>He held his staff at his side</quote>
            </cit>
        </cit>
    <bibl>(1225 Lay. Brut 30784 [MED])</bibl>
</cit>

<cit type="etymon" xml:id="at-1225-b" prev="#at-1225-a" next="#at-c1300">
    <oRef xml:lang="enm"><seg>bi</seg><pc>-</pc>siden</oRef>
    <gloss>in addition</gloss>
    <gramGrp>
        <pos>conjunctiveAdverbial</pos>
    </gramGrp>
    <cit type="component" corresp="#e3s1">
        <gramGrp>
            <pos>preposition</pos>
        </gramGrp>
```

---

⁵¹In this example, the adverbial component of the etymon, the additional feature of *prefix,* which would be specified as the entry type within a synchronic entry, could be labeled within an @ana attribute, e.g. <cit type="component" corresp="#e2s1" ana="#prefix">.

```xml
        </cit>
        <cit type="component" corresp="#e3s2">
            <gramGrp>
                <pos>nominalAdposition</pos>
            </gramGrp>
        </cit>
        <cit type="component" corresp="#e3s3">
            <gramGrp>
                <case>dative</case>
            </gramGrp>
        </cit>
        <cit type="attestation" xml:lang="ang">
            <quote>Heo letten forð<oRef>bi-siden</oRef> an oþer folc riden, ten
þusend kempen</quote>
            <cit type="translation" xml:lang="en">
                <quote>They sent another army forth in addition, 10,000
warriors</quote>
            </cit>
        </cit>
        <bibl>(1225 Lay Brut 5498 [MED])</bibl>
</cit>[52]
```

**Stage I:** Adverbial of extension/verbal adverbs

A major feature of interest in adverbials of extension is where they occur at the end of a clause;
in these cases, the adverbs are often an oblique syntactic argument. This is the case in the section
of the example below in which the tag 'extensionAdverb' is used in the source for adverbials of
extension[53].

- prefix 'be-' now fused onto 'side'
- adverb oblique argument, labeled as value of @ana in the <oRef> in the attestation

```xml
<cit type="etymon" xml:id="at-1514-1518" prev="#at-1450" next="#at-1535-1543">
    <date notBefore="1514" notAfter="1518"/>
    <oRef xml:lang="en">beside</oRef>
    <gramGrp>
        <pos>extentionAdverb</pos>
    </gramGrp>
    <cit type="attestation"><!-- early modern english-->
        <quote>In whiche albeit thei ment as muche honor to hys grace as wealthe
to al the realm <oRef ana="#Oblq">beside</oRef>, yet were they not sure howe hys grace
woulde take it,</quote>
    </cit>
    <bibl>(1514-18 More, History of Richard III, p. 78)</bibl>
</cit>

<cit type="etymon" xml:id="at-1535-1543" prev="#at-1514-1518" next="#at-1567">
    <date notBefore="1535" notAfter="1543"/>
    <oRef xml:lang="en">beside</oRef>
    <gramGrp>
        <pos>extentionAdverb</pos>
    </gramGrp>
    <cit type="attestation">
        <quote>The toune of Chester is chiefly one streate of very meane building
yn lenght: ther is <oRef>beside</oRef> a smaul streat or 2. about the chirch; that is
collegiatid, .... </quote>
    </cit>
    <bibl>(1535-43 Leland, Itinerary I p. 74 [HC])</bibl>
</cit>
```

---

[52] One etymon from the source has been skipped in this example: 'besyde' in the prepositional sense attested ca.
1450
[53] 'Extension adverbials' are also referred to by the Traugott as 'verbal adverbials'

**Stage II**:Sentential Adverb

One of the two contextual conditions in which sentential adverbials (sententialAdverb) occur is that they need to immediately follow a complementizer. This is represented in the data by the encoding of the relevant item within the <seg> element and the @ana attribute is used to label it with the needed data category (complementizer).

```xml
<cit type="etymon" xml:id="at-1552-1563" prev="#at-1535-1543" next="#at-1554">
    <date notBefore="1552" notAfter="1563"/>
    <oRef xml:lang="emodeng">besides</oRef>
    <gramGrp>
        <pos>sententialAdverb</pos>
    </gramGrp>
    <cit type="attestation">
        <quote>…when the endisknowen,allwilturnetoaiape ['trick,deceit],
Tolde he not you <seg ana="#Complimentizer">that</seg><oRef>besides</oRef> she stole
your Cocke that tyde?</quote>
    </cit>
    <note>The adverbial of extension which signals the extension of a list of
referents as per (sense in example "at-#at-1535-1543") is presumed to be the source of
the clause-initial, focused sentence adverbial, which extends the propositional
content with additional, non-central material:</note>
    <bibl>(1552-63 Gammer Gurton, p. 61 [HC])</bibl>
</cit>
```

**Stage III**: Discourse Marker/Particle

In this next stage, 'besides' serves to refocus the attention on the purposes of the discourse,andoccurs on left periphery of the sentence structure serving a pragmatic function. As in the previous example, the context is labeled using <seg> with the attribute @ana

```xml
<cit type="etymon" xml:id="at-1554" prev="#at-1552-1563" next="#stage3-26a">
    <date when="1554"/>
    <oRef xml:lang="emodeng">besides</oRef>
    <gramGrp>
        <pos>discourseMarker</pos>
    </gramGrp>
    <cit type="attestation">
        <quote>...<seg ana="#LPeriph">And <oRef>besides</oRef></seg>, it is very
unlike, that </quote>
    </cit>
    <bibl>(bef. 1554 Trial Throckmorton I,66.C1 [HC])</bibl>
</cit>⁵⁴
```

In these examples from 1619 and 1872 the usage of 'beside' extends the discourse with after thoughts:

```xml
<cit type="etymon" xml:id="at-1619" prev="#at-1554" next="#at-1872">
    <oRef xml:lang="emodeng">beside</oRef>
    <gramGrp>
        <pos>discourseMarker</pos>
    </gramGrp>
    <cit type="attestation">
        <quote>..and<oRef>beside</oRef>,   my   complexion   is   so   blacke,
that..</quote>
    </cit>
    <bibl>(1619 Deloney, Jack of Newbury, p.70 [HC])</bibl>
</cit>

<cit type="etymon" xml:id="at-1872" prev="#at-1619">
    <oRef xml:lang="en">besides</oRef>
    <gramGrp>
        <pos>discourseMarker</pos>
    </gramGrp>
```

---

⁵⁴etymon skipped from source: 'besides' in sense of ExtentionAdverb attested 1567

```xml
<cit type="attestation">
        <quote>The whooping cough seems to be a providential arrangement to force
you to come, as the expense will be little greater than going anywhere else;
<oRef>besides</oRef> if you put a trusty female at Ravenscroft… <quote>
        </cit>
        <bibl>(1872 Amberley Ltrs, p. 513 [CLME])</bibl>
</cit>
```

Finally the source of the etymological data is cited as a child element of <etym>:

```xml
        <bibl>(Traugott, 1995)</bibl>
      </etym>
    </sense>
  </entry>
```

This example is obviously an extremely complex yet informative representation of the history of the lexical item in question (English: *besides*). Such information as that in the source paper, while important within the field of Linguistics in the furthering the understanding of such central etymological processes such as grammaticalization, is never modeled in a reusable markup format such as XML, nor is it included in any kind of dictionaries, etymological or traditional. It is thus, a major untapped resource of highly advanced etymological data models. Establishing a practice in which this kind of markup is done may help linguists and lexicographers begin to more regularly make maximum use (and re-use) of the research being done.

## 11. Problematic and Unresolved Issues

For the issues regarded as the most fundamentally important to creating a dynamic and sustainable model for both etymology and general lexicographic markup in TEI, we have submitted formal requests for changes to the TEI GitHub, with more likely to be necessary moving forward.While this work represents a large step in the right direction for those looking for means of representing etymological information, there are still a number of unresolved issues that will need to be addressed. These remaining issues pertain to: (i) expanding the types of etymological information and refining the representation of the processes and their features which are covered; and (ii) the need for continued progress in a number of issues within the body of international standards on which lexical markup relies.

Some examples of issues from (i) are as follows:

- encoding onomasiological etymological information, which groups and represents the converging and diverging histories of related forms across multiple related languages;
- related is the markup of existing non-semasiological etymological dictionaries which can be extremely long and which can be organized in extremely complex ways;
- sense shifts not involving metaphor or metonymy, but subtle, often usage- and context-based patterns such as: pejoration-amelioration, narrowing-widening, etc.
- changes to a lexical item which stem from corresponding changes and functional linguistic pressures occurring within other parts of the lexicon, such as *semantic bleaching;*

Examples of issues from (ii) are:

- the inventory of categories in ISOcat[55] should be expanded to include well established etymological processes, such as metaphor, metonymy, etc., as well as other features not specific to etymology;

---

[55]ISOCat is currently stalled with two initiatives from ISO TC 37 and Clarin taking up the legacy. We shall be working in making the two converge again.

Several significant issues regarding language identification are as follows:

- the need for continued expansion of the inventory of the IANA registry, probably in conjunction with the maintenance of future versions of ISO 639;
- the need for more granularity in the means by which @xml:lang is specified without using the private tag ("-x-") (see BCP 47); currently most specific one can be in labeling language varieties while remaining interoperable is the specification of the following information: (language-country-region); related to this larger issue of language tagging are the following problems for etymological markup:
- there are no existing codes within any body of standards for historical places such as Gaul, Carthage, etc.;
- there is no way to label when the language data is from an intermediate, transitional period for which there is neither IANA entries, or even a commonly used term within the body of literature

## 12. Conclusion

In this paper we have proposed a number of markup scenarios for encoding different types of etymological information within TEI dictionaries drawing from both traditional lexicographic practice and from analytical approaches from functional and cognitive linguistics. Our examples have shown a number of cases in which, in order to represent the lexical data as accurately as possible, it has been necessary to alter the content models of certain elements and attributes in a way we think should actually be considered as necessary evolutions for the TEI guidelines in the future. Relevant to both etymological and synchronic lexical markup, we have also touched upon how encoders can make use of linked open data URI's as a means of linking the sense and semantic domain(s) of an item to multilingual knowledge bases. Future work that remains to be addressed is the issue of encoding of onomasiological etymological data such as that found in dialectal and traditional etymological dictionaries.

2005, Vancouver, Canada.

# Biographies

## Jack Bowers

Jack Bowers works as a research assistant at the Austrian Academy of Sciences (ÖAW)- Austrian Center for Digital Humanities (ACDH).

He is the primary curator of the DBÖ (Datenbank der bairischen Mundarten in Österreich) corpus and is working on cleaning, converting and enhancement of the lexical and metadata in order to facilitate reuse, and to bring it in line with international standards and best practice.

With a background in cognitive and functional approaches to all levels of linguistics and their interfaces (semantics, morpho-syntax, phonetics, phonology, etymology, etc.) a major area of focus is on applying linguistic expertise to best represent and integrate lexical data (spoken or textual) with metadata and semantic knowledge/information resources. Also of particular interest are working towards interoperability between standards for lexical markup (TEI, LMF, ONTOLEX, TBX) and in the emerging prospects offered by semantic web/ontological resources in the integration of human knowledge across academic and scientific fields.

Jack has a B.A. in History and French from San Francisco State University (2009) and an M.A. in Linguistics and a certificate in Computational Linguistics at San Jose State University (2012). He is a member of the DARIAH-GiST project which works to promote the use of digital standards in the humanities and became involved in the digital humanities while working on his documentation/multimedia resource/corpus creation of the Mixtepec-Mixtec language variety (spoken Juxtlahuaca district, Oaxaca Mexico) (2011-).

## Laurent Romary

Laurent Romary is Directeur de Recherche at Inria, France, director general of DARIAH. and guest scientist at the Academy of Sciences in Berlin. He carries out research on the modeling of semi-structured documents, with a specific emphasis on texts and linguistic resources. He received a PhD degree in computational linguistics in 1989 and his Habilitation in 1999. During several years he launched and directed the Langue et Dialogue team at Loria in Nancy, France and participated in several national and international projects related to the representation and dissemination of language resources and on man-machine interaction. He is the chairman of ISO committee TC 37 and has been member (2001-2007) then chair (2008- 2011) of the TEI (Text Encoding Initiative) council. Beyond his research activities, he has been responsible for defining and implementing the scientific information and open access policy of major research institutions in Europe, namely CNRS, Max Planck Society and Inria. . He currently contributes to the establishment and coordination of the European Dariah infrastructure.